\documentclass[conference]{IEEEtran}
\IEEEoverridecommandlockouts
\usepackage{cite}
\usepackage{amsmath,amssymb,amsfonts}
\usepackage{algorithmic}
\usepackage{graphicx}
\usepackage{textcomp}
\usepackage{xcolor}
\usepackage{subcaption}
\usepackage{hyperref}
\def\BibTeX{{\rm B\kern-.05em{\sc i\kern-.025em b}\kern-.08em
    T\kern-.1667em\lower.7ex\hbox{E}\kern-.125emX}}
\usepackage{draftwatermark}
    
\usepackage[markup=bfit, deletedmarkup=sout, authormarkup=brackets]{changes}
\definechangesauthor[name={Filipe}, color=blue]{FG}
\definechangesauthor[name={Matej}, color=orange]{MH}
\definechangesauthor[name={Matthias}, color=green]{MR}

\begin{document}
\SetWatermarkAngle{0}
\SetWatermarkColor{black}
\SetWatermarkLightness{0.5}
\SetWatermarkFontSize{10pt}
% \SetWatermarkScale
% \SetWatermarkHorCenter
\SetWatermarkVerCenter{20pt}
\SetWatermarkText{\parbox{30cm}{%
\centering This is the author final version of the manuscript accepted for publication in\\
\centering 2020 Joint IEEE 10th International Conference on Development and Learning and Epigenetic Robotics (ICDL-EpiRob), (C) IEEE}}

\title{Active exploration for body model learning through self-touch on a humanoid robot with artificial skin\\
}

\author{\IEEEauthorblockN{Filipe Gama\IEEEauthorrefmark{1}, Maksym Shcherban\IEEEauthorrefmark{1}, Matthias Rolf\IEEEauthorrefmark{2}, Matej Hoffmann\IEEEauthorrefmark{1}}
\IEEEauthorblockA{\IEEEauthorrefmark{1}\textit{Department of Cybernetics, Faculty of Electrical Engineering}, \textit{Czech Technical University in Prague}, \\
Prague, Czech Republic \\
filipe.gama@fel.cvut.cz, matej.hoffmann@fel.cvut.cz}
\IEEEauthorblockA{\IEEEauthorrefmark{2}\textit{School of Engineering, Computing and Mathematics, Oxford Brookes University} \\
%\textit{Faculty of Technology, Design and Environment, Oxford Brookes University}\\
Oxford, United Kingdom \\
mrolf@brookes.ac.uk}
}

\maketitle

\begin{abstract}
The mechanisms of infant development are far from understood. Learning about one's own body is likely a foundation for subsequent development. Here we look specifically at the problem of how spontaneous touches to the body in early infancy may give rise to first body models and bootstrap further development such as  reaching competence. Unlike visually elicited reaching, reaching to own body requires connections of the tactile and motor space only, bypassing vision. Still, the problems of high dimensionality and redundancy of the motor system persist. In this work, we present an embodied computational model on a simulated humanoid robot with artificial sensitive skin on large areas of its body. The robot should autonomously develop the capacity to reach for every tactile sensor on its body. To do this efficiently, we employ the computational framework of intrinsic motivations and variants of goal babbling---as opposed to motor babbling---that prove to make the exploration process faster and alleviate the ill-posedness of learning inverse kinematics. Based on our results, we discuss the next steps in relation to infant studies: what information will be necessary to further ground this computational model in behavioral data.
\end{abstract}
%This document is a model and instructions for \LaTeX.
%This and the IEEEtran.cls file define the components of your paper [title, text, heads, etc.]. *CRITICAL: Do Not Use Symbols, Special Characters, Footnotes, or Math in Paper Title or Abstract.


\section{Introduction}
Touch is the first sense to emerge in the fetus \cite{Bradley1975}. Fetuses perform local movements directed to areas of the body most sensitive to touch: the face (the mouth in particular), but also for example soles of feet \cite[p.~ 113-114]{Piontelli2015}. Later, from 26 to 28 weeks of gestational age, they also use the back of the hands to touch as well as touch other body areas like thighs, legs, and knees \cite[p.~ 29-30]{Piontelli2015}. In addition, from 19 weeks, fetuses anticipate the hand-to-mouth movements \cite{Myowa2006} (the mouth opens prior to contact) and from 22 weeks, the movements seem to show the recognizable form of intentional actions, with kinematic patterns that depend on the goal of the action (toward mouth vs. toward eyes) \cite{Zoia2007}. Birth obviously brings about a major disruption of the equilibrium that was reached in the womb: the constrained aquatic environment is suddenly replaced by an aerial one, with gravity playing a major part. Nevertheless, hand-mouth coordination continues to develop after birth (e.g.,~\cite{Rochat1993}). Also, Thomas et al.~\cite{Thomas2015}, biweekly recording resting alert infants from birth to 6 months of age, show that infants do frequently touch their bodies, with a rostro-caudal progression as they grow older: Head and trunk contacts are more frequent in the beginning, followed by more caudal body locations including hips, then legs, and eventually the feet. DiMercurio et al.~\cite{DiMercurio2018}, following infants from 3 to 9 weeks after birth, found no consistent differences in the rate of touch between head and trunk. In summary, infants acquire ample experience of touching their body. The question remains what drives this behavior and how this experience is catalogued and used to develop first tactile-proprioceptive-motor models of the body. The dynamic brain development in this period has to be considered as well (see \cite{Tau2010}; \cite{hoffmann_selfTouch_2017} for an account focusing specifically on the somatosensory areas).

Are the touches to the body spontaneous or systematic? If there is a particular structure---which seems to be the case \cite{DiMercurio2018,Thomas2015}---what drives this developmental progression? Piaget~\cite{Piaget1952} theorized that in newborns, action and perception as well as the ``spaces'' of individual sensory modalities are separated. Until the connections (a ``model'') are established, infants explore their environment (and their body) randomly. However, there is now evidence that the modalities are already connected early after birth (e.g., \cite{vanDerMeer1995} for the visual and motor). 
Also, there is empirical evidence that infants perform goal-directed action right from the outset of motor learning---reviewed in \cite{Bertenthal1996}. Specifically related to body exploration, Rochat~\cite{Rochat1998} writes: ``By 2-3 months, infants engage in exploration of their own body as it moves and acts in the environment. They babble and touch their own body, attracted and actively involved in investigating the rich intermodal redundancies, temporal contingencies, and spatial congruence of self-perception.'' 
%``young infants' propensity to engage in self-perception and systematic exploration of the perceptual consequences of their own action plays an important role in the intermodal calibration of the body and is probably at the origin of an early sense of self: the ecological self. '' \cite{Rochat1998}

The goal of this work is to operationalize these observations and hypotheses using a synthetic approach, ``understanding by building'', by developing embodied computational models of the phenomenon---typical for cognitive developmental robotics \cite{asada_2009,hoffmann_pfeifer_2018}. Exploration through random movements---often dubbed \textit{body babbling} \cite{Meltzoff1997} or \textit{motor babbling}---has been employed in different models (e.g., the ``endogenous random generator'' in \cite{Bullock1993}). However, faced with the dimensionality of the motor and sensory spaces, trying out all the possible combinations of motor commands and observing their consequences is hugely inefficient. For example, most motor commands generate movements that do not result in any contact with the body and hence do not generate useful experience to learn the motor-tactile contingencies. Therefore, we employ two key ideas that help the agent to channel the exploration in the right direction. First, the agent should monitor its learning efficiency---the gain in its knowledge or competence to achieve specific goals---and focus the exploration on regions of the search space that are currently most promising. This is exemplified by the computational frameworks dealing with intrinsic motivation (or artificial curiosity) \cite{Schmidhuber1991,OudeyerKaplan2007,baranes_2013,Baldassarre2013}. Second, the agent should focus the exploration on the goal space rather than the motor space. The goal space---the skin on the body in our case---may be lower-dimensional and it is here where the ``interest'' of the agent lies. If it does babble, it should thus do \textit{goal babbling} \cite{Rolf2010} rather than \textit{motor babbling}.  

This article is structured as follows. After reviewing related work in the next section, Section~\ref{sec:materials} presents the robot simulator and the exploration framework. After experimental results (Section~\ref{sec:results}), we summarize them (Section~\ref{sec:conclusion}) and discuss their implications and future work (Section~\ref{sec:discussion}). An accompanying video is available here: \url{https://youtu.be/Zb87uTFnQZE}.

\section{Related work}
\label{sec:rel_work}
Our focus are ``mechanisms that drive a learning agent to perform different activities for their own sake, without requiring any external reward'' \cite{baranes_2013}. This phenomenon has been articulated in psychology as intrinsic vs. extrinsic motivation---\cite{Ryan2000} provides an overview. Oudeyer and Kaplan~\cite{OudeyerKaplan2007} strive to clarify the terms of internal/intrinsic and external/extrinsic rewards and present a computational perspective as well as relationship to other computational frameworks such as reinforcement learning. As briefly outlined above, there are two key aspects of efficient exploration: (i) monitoring learning progress and (ii) focusing on the ``goal space''. The former has been addressed by a number of frameworks that can be classified as \textit{knowledge-based} \cite{OudeyerKaplan2007}. The latter aspect has been addressed by the path-based goal babbling approach of Rolf et al.~\cite{Rolf2010} or by other, \textit{competence-based} approaches in which the agent self-generates goals that it tries to accomplish. The idea is best illustrated on the example of learning to reach, or, learning inverse kinematics. The motor system is known for its redundancy: there are multiple ways of reaching to a specific point in space. Knowledge-based approaches that monitor learning progress but are confined to the motor space (e.g., \cite{Baranes2009}) will discover multiple solutions to the same goal, which can often be considered inefficient. Moreover, the space of solutions in the joint space (motor space) is not convex: averaging between them will often result in wrong configurations. Rolf et al.~\cite{Rolf2010} analyze this and develop a solution, goal babbling, that deals with this problem: by exploring in the goal space, the agent is not ``motivated'' to look for alternative solutions. Further, following continuous  paths through the goal space allows to circumvent the issue of non convex solutions \cite{Rolf2010}. This architecture has been also used to model the U-shaped curve typical of infant development \cite{Narioka2015}. Baranes and Oudeyer extended their R-IAC (Robust Intelligent Adaptive Curiosity) architecture \cite{baranes_2013} to Self-Adaptive Goal Generation Robust Intelligent Adaptive Curiosity (SAGG-RIAC)---a competence-based strategy---that also handles learning inverse kinematics in redundant manipulators. Our work is employing the computational framework of \cite{baranes_2013}, as embedded in the \textit{Explauto} library \cite{Moulin2014}.

Learning to discover the surface of the body---a 2-dimensional skin surface embedded in the 3-dimensional world and moving together with the body parts---is similar to the problem of learning inverse kinematics that is a typical showcase for many of the intrinsic motivation frameworks (e.g., \cite{Rolf2010,baranes_2013}). The motor space or joint space is identical; the goal space, or \textit{observation space}, also \textit{interest space}, is different: for learning inverse kinematics, these are 3D Cartesian coordinates of the end effector (the infant hand, say). For the body space, either skin activations or spatial coordinates are candidate representations, which will be explained in detail in Section~\ref{sec:materials}). The work of Kuniyoshi, Mori and colleagues (e.g., \cite{Mori2010,Yamada2016}) on the fetus simulator is complementary to this work, addressing prenatal development and focusing on a lower level: first tactile-motor interactions are emerging from the musculoskeletal body model coupled to spinal and simple subcortical or cortical circuitry. In comparison, the present study focuses on how guided exploration on a higher level of abstraction can give rise to efficient body exploration.  

The work most related to ours is that of Mannella et al.~\cite{mannella_2018} who specifically target the body (skin surface) as the exploration target. Their architecture is rather complex compared to ours, consisting of Goal generator, Goal selector, Motor controller, and Predictor. The simulated agent, however, is quite simple, consisting of two arms in 2D with three degrees of freedom (DoF) each, and a ``skin'' emulated using 30 Gaussian receptive fields in a 1D topology. 
%has two arms and a torso concurrently - arms can touch each other
% our skin is more realistic and also more challenging with actual contact sensors that are binary!?
The motor controller is also highly complex, composed of a dynamic-reservoir recurrent neural network, a random generator, and associative memory. The ``skin receptors'' are phasic, as they respond to changes rather than sustained values. These changes are then relayed into a self-organizing neural map (SOM) that ``clusters'' them. Compared to this, our architecture is much simpler. The motor space consists simply of the robot joint space. That is, only the final configurations/postures matter---motor overlaps with proprioceptive---and the actual movement production is sidestepped. 
%This is different to \cite{mannella_2018} or \cite{baranes_2013}, the latter employing also a kind of differential kinematics---changes of joint angles. 

\section{Materials and Methods}
\label{sec:materials}
This section provides an overview of the robot simulator and the exploration framework. 

\subsection{Nao humanoid robot with artificial skin}
\label{sec:nao_with_skin}
\begin{figure}[!ht]
\centering
\begin{subfigure}{.24\textwidth}
  \centering \includegraphics[scale=1, width=1\linewidth,trim = 0cm 5cm 0cm 3cm, clip=true]{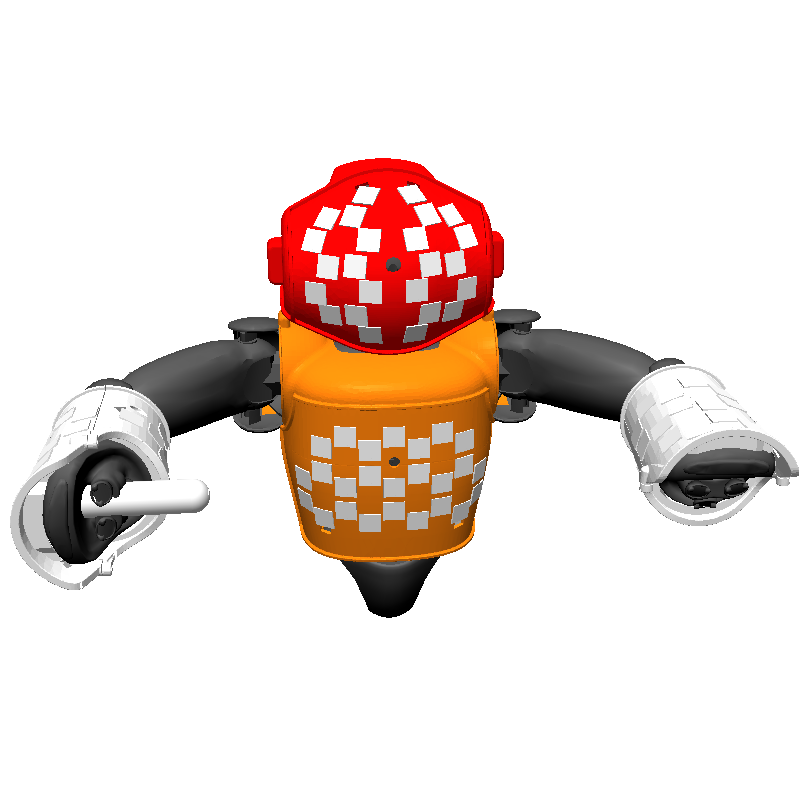}
\end{subfigure}%
\begin{subfigure}{.24\textwidth}
  \centering \includegraphics[scale=1, width=1\linewidth,trim = 0cm 5cm 0cm 3cm, clip=true]{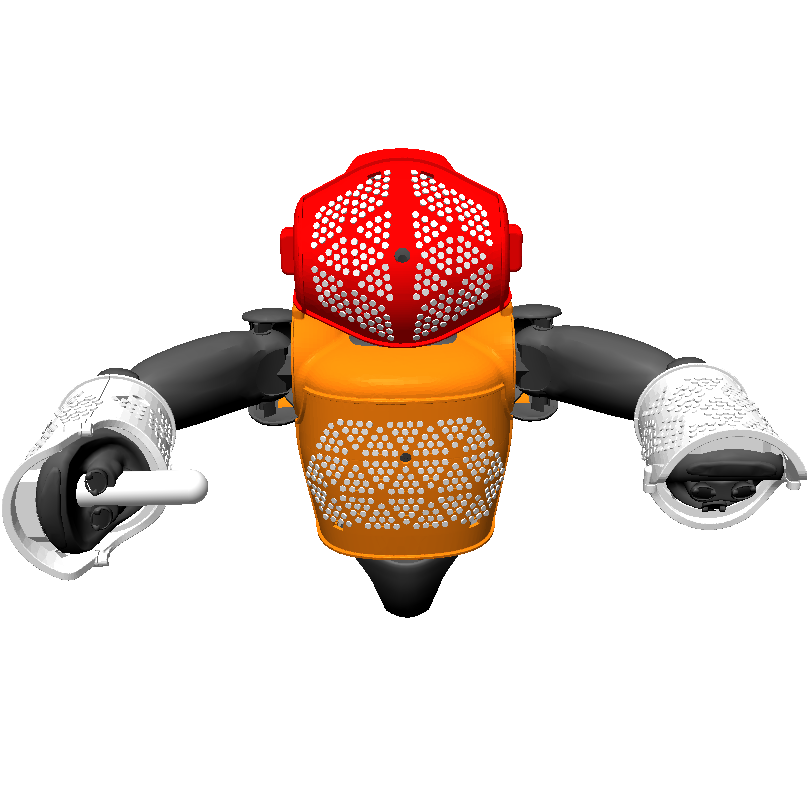}
\end{subfigure}
\caption{Nao robot model in Gazebo. (Left) Low-resolution skin. (Right) High-resolution skin.}
\label{fig:naogazebo}
\end{figure}

The experimental platform was a simulated Nao humanoid robot in Gazebo 9. 
%Different aspects of the simulation are implemented in separate ROS nodes which communicate through ROS services and topics.
The model used is a variant of the publicly available \textit{naov40} URDF model, modified to add the parts hosting tactile/pressure sensors (``skin'') using the Gazebo ContactSensor plugin. There are two variants of the skin: (i) low-resolution (Fig.~\ref{fig:naogazebo} left) and (ii) high-resolution (Fig.~\ref{fig:naogazebo} right). The latter mimics the physical Nao robot available at CTU, uniquely equipped with ``iCub skin sensors'' \cite{Maiolino2013}. Low-resolution skin has 25, 24 and 27 sensors for the torso, the head, and each wrist respectively; high-resolution skin has 250, 240 and 270 tactile sensors on the same body parts.

The code of the simulator is available at \cite{nao_gazebo}.
%The robot fingers cannot be made stiff and hence cannot reliably generate skin activation. 
A cylindrical ``pen'' tool with spherical endpoint was attached to the robot's wrist to act as a finger and facilitate localized touch. 
%, replacing the original fingers that could not be controlled and used properly to generate touches.
%where:
%\begin{itemize}
 % \item The base of the robot's torso was fixed in space
  %\item The legs, fingers, gripper, cameras and sonar sensors were removed for simplification as they are not used
  %\item The plastic casings that contain the artificial skin on the physical robot were added on the model's torso, head and wrists
  %\item Taxels were placed on the casings, and touch feedback was enabled with the Gazebo ContactSensor plugin  adapted to support Gazebo 9 API.
  %\item A cylindrical pen tool with spherical endpoint was attached to the robot's wrist to act as a finger and facilitate touch of a single taxel, replacing the original fingers that could not be controlled and used properly to generate touches.
%\end{itemize}
 More details can be found in \cite{Shcherban2019}.

%PARTS CAN BE TAKEN FROM MAX'S THESIS. (LINK TO SOURCE BELOW)
% link to source files: \url{https://drive.google.com/open?id=1kTBduRRnu1DpHzzq7VkXCWeM_twiA3vO}

\subsection{Explauto library}
\label{sec:explauto}
Explauto (\cite{Moulin2014}, \url{https://github.com/flowersteam/explauto}) is a framework for implementation and benchmarking of sensorimotor learning algorithms, with a specific focus on intrinsic motivation: monitoring learning progress in motor or sensory (goal) spaces.
%It provides sensorimotor and interest models that can be used to generate goals (goal babbling) in the observation space for learning forward and inverse models using predictions.
The \textit{action space} \textit{Q}, represents all possible actions (e.g., joint configuration) of the robot. An action $\textit{q}\in\textit{Q}$  generates an outcome $\textit{x}\in\textit{X}$ in the \textit{observation space} \textit{X}. 
%\mh{The two sentences above are already specific to our case and should me (re)moved / integrated into Sec.~\ref{sec:action_and_obs_spaces}.} \fg{But they introduce the terms for the next sentence to explain what is the \textit{interest space} used/defined by Explauto. It is also not that specific, since this is how Explauto defines the spaces used for motor and goal babbling} Explauto exploration strategy is performed on an \textit{interest space}, which is the action space for motor babbling strategies or the observation space for goal babbling strategies.
%Explauto uses the notion of \textit{lazy learning}: training data is processed only when a query is asked.
During exploration, a database is constructed, with every entry being a tuple: $(\textit{q},\textit{x})$.
%In our experiments, as the focus is on learning from touch, only actions that generated a touch feedback are added to the database.\mh{Again, last sentence should probably not go here.}

\subsection{Action and observation spaces}
\label{sec:action_and_obs_spaces}

Only the upper body of the Nao robot, which hosts the artificial skin, is used. The robot uses one of its arms to touch either the torso or the face. 
%Only the joints on each arm (5 per arm) and the neck (2) are used, making it 12 in total, and ranging from 5 to 7 DoF depending on the experiment. Only one hand at a time was used during experiments to avoid frequent wrist collisions that block the robot from reaching the skin on the torso or the head. These collisions were more frequent than they should be on a regular Nao robot because of the casings added on the wrist for the artificial skin, making them wider. 
Its \textit{action space} is the robot joint space, with five degrees of freedom per arm and two DoF on the neck. To touch the torso, only the arm is used, hence $\textit{Q} \subseteq \mathbb{R}^{5}$; to touch the head, the neck joints also contribute: $\textit{Q} \subseteq \mathbb{R}^{7}$. 
The joint ranges can be found in \cite{Shcherban2019}.

%\mh{Please put a well-structured, perhaps using bullet points, and formal (mathematical) description of the spaces.}
%\fg{I'm unsure about the bullet points, isn't the paragraph enough? As for the formal description, since they are quite well-known 'shapes' (plan and cylinder), and the geometrical projections are quite well-known, I'm not sure if we need to define them much more formally. I changed a few sentences to make the transformations a bit more clear, but unless I'm missing something, the main "changing part" would be the radius of the cylinder, and I don't think it is interesting enough to mention, especially if we lack space.}
%\mh{The skin projections should go here. They define a \textit{metric} on the space which I guess is something we should discuss.}

\begin{figure}[!ht]
\centering
\begin{subfigure}{.17\textwidth}
  \centering \includegraphics[scale=1, width=1\linewidth]{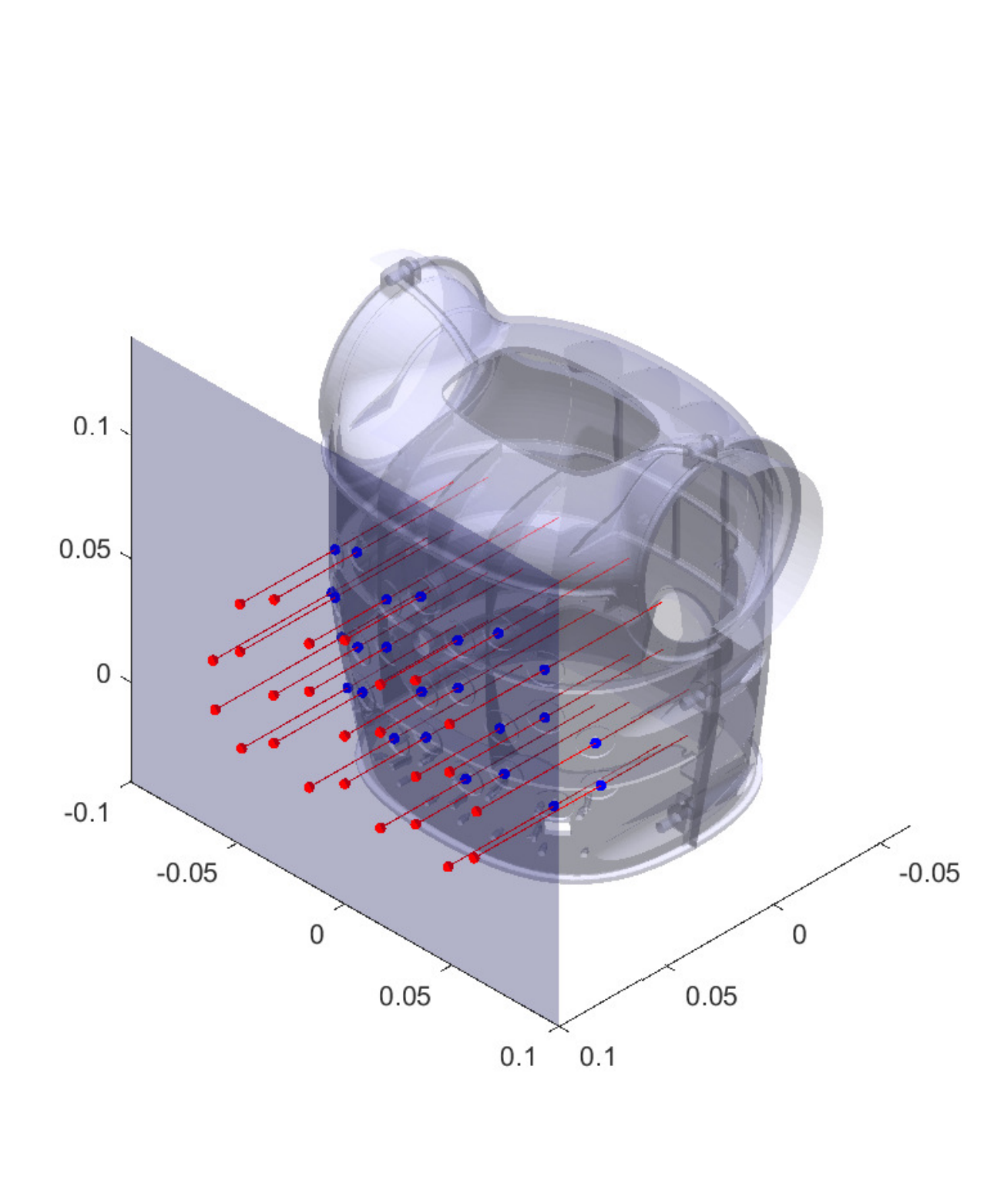}
  (a)
\end{subfigure}%
\begin{subfigure}{.17\textwidth}
  \centering \includegraphics[scale=1, width=1\linewidth]{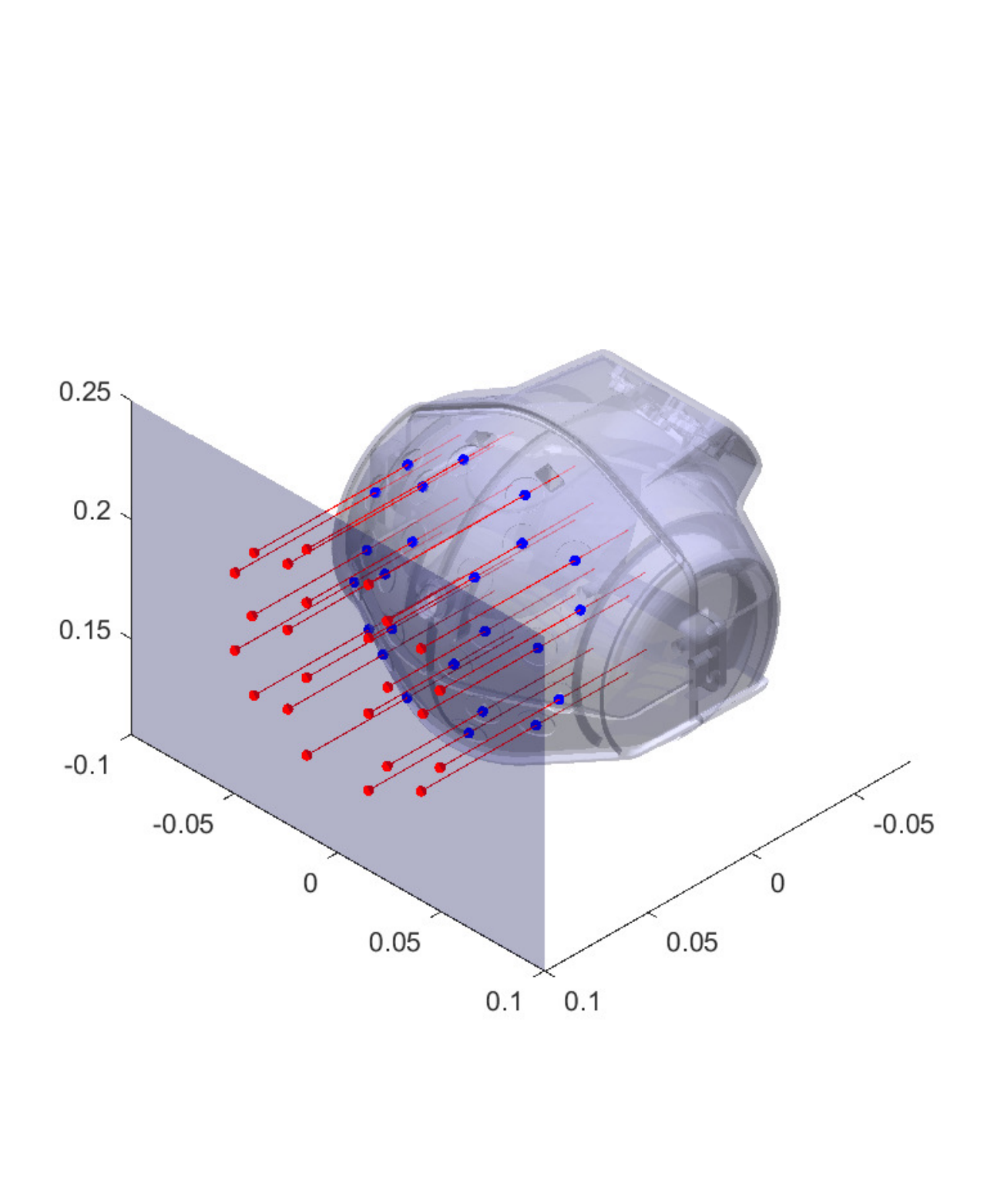}
  (b)
\end{subfigure}

\begin{subfigure}{.17\textwidth}
  \centering \includegraphics[scale=1, width=1\linewidth]{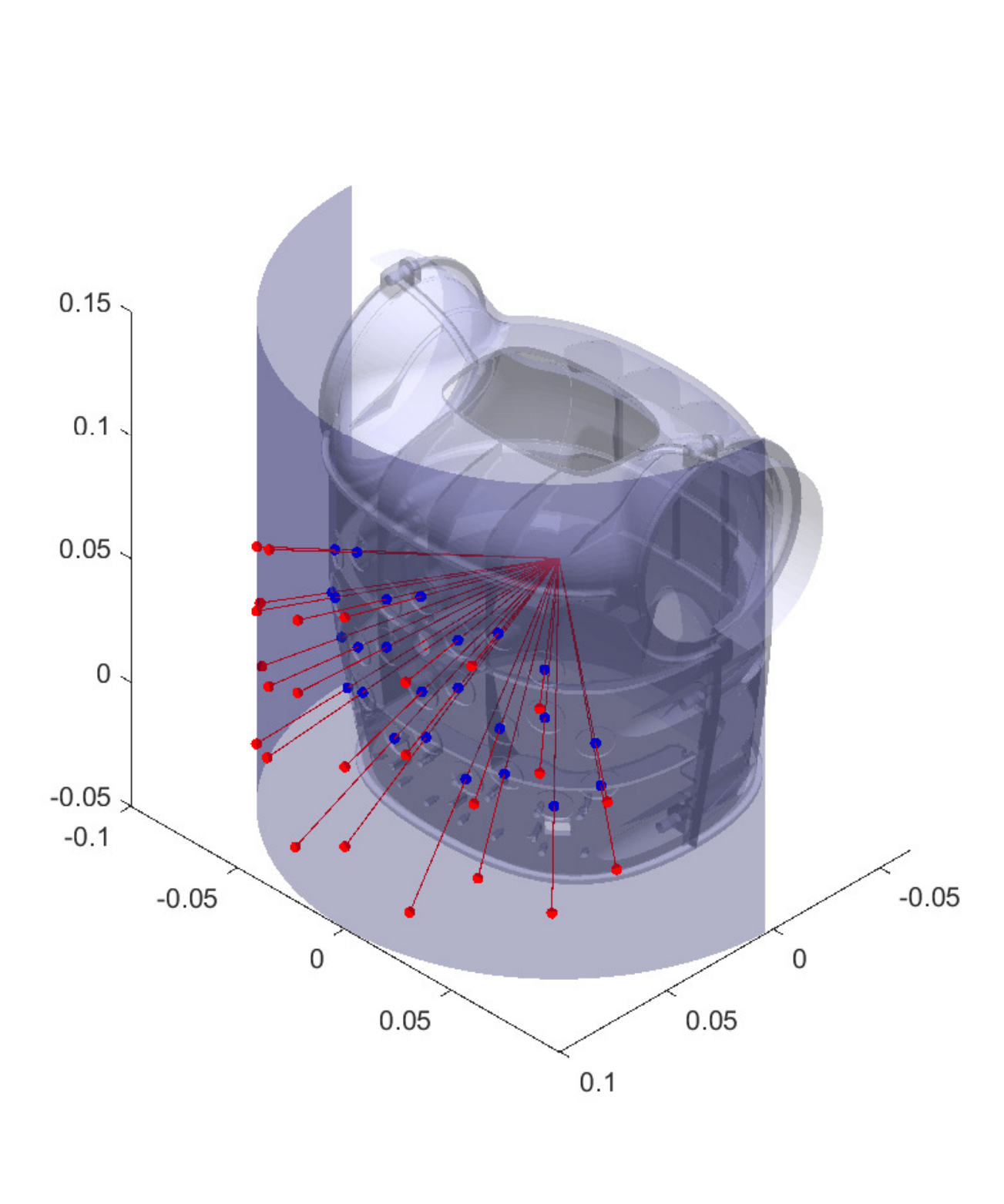}
  (c)
\end{subfigure}%
\begin{subfigure}{.17\textwidth}
  \centering \includegraphics[scale=1, width=1\linewidth]{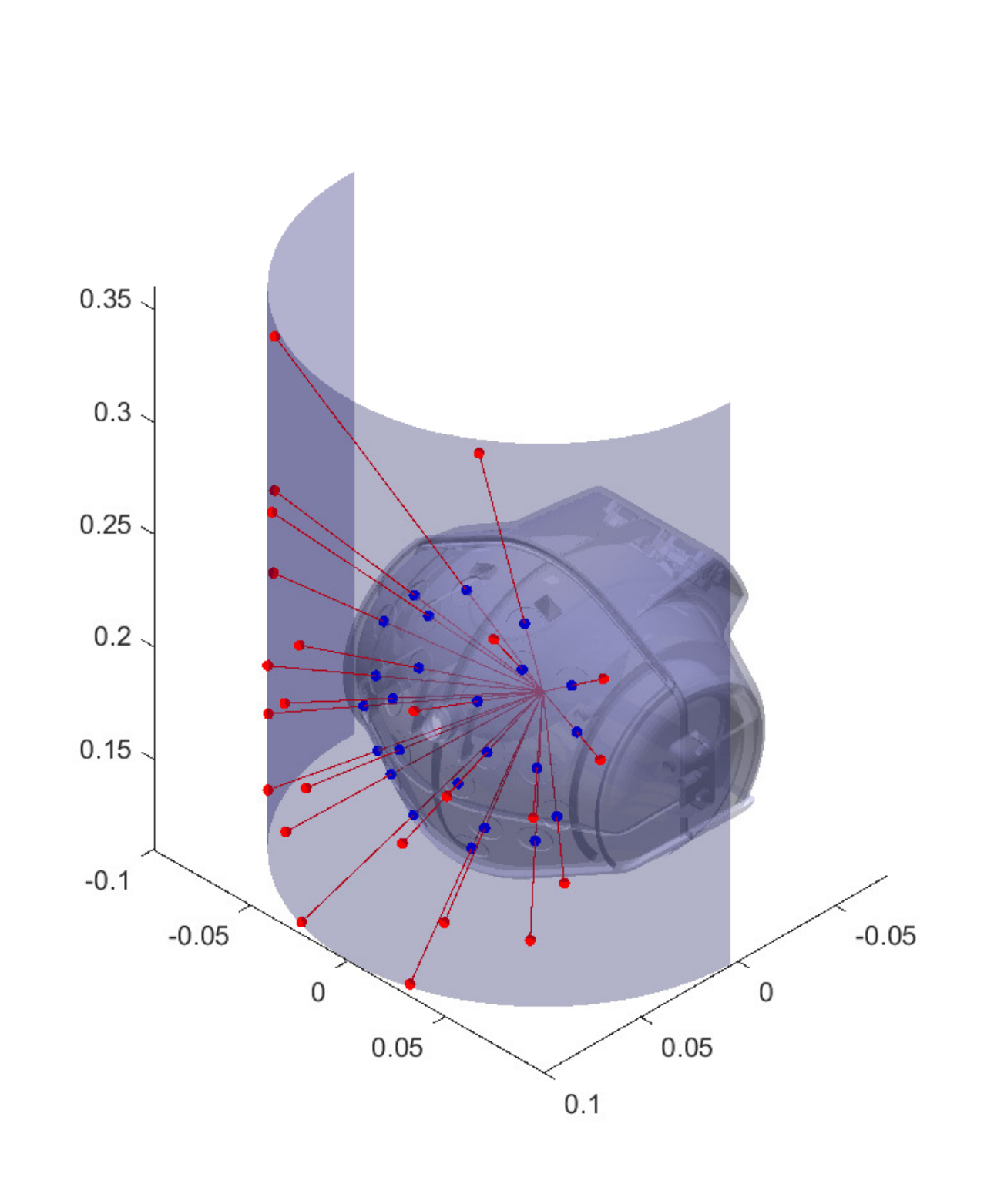}
  (d)
\end{subfigure}
\caption{Projection of taxels' coordinates. Top row: parallel planar projections used for low-resolution skin on (a) torso and (b) head. Bottom row: central cylindrical projection for high-resolution skin on (c) torso and (d) head.}
\label{fig:plannar-proj}
\end{figure}

The \textit{observation space} is the robot skin-activation generated when the robot contacts its torso or face with its arm. This is a discrete space of individual taxels and their activation. Taxel activation is binary: either a taxel is activated or it is not. For the exploration methods considered here, a distance \textit{metric} on this space is needed. Connecting neighboring taxels by edges and acquiring distances from an incidence graph would be a possibility. To aid the computational exploration framework, we formulated a continuous metric on the observation space.
A two-dimensional observation space centered at its origin is created for each body part, using a projection of the taxels to obtain their coordinates on the new space. A simple planar parallel projection was used for low-resolution skin (Fig.~\ref{fig:plannar-proj}, a and b), while a central projection on a cylindrical surface was used for high-resolution skin (Fig.~\ref{fig:plannar-proj}, c and d), representing the shape of the body parts more accurately. For each body part, we thus have $\textit{X} \subseteq \mathbb{R}^2$.
%Because the sub-spaces for each body part are two-dimensional, the observation space can be defined as $\textit{X} \subseteq {\!R}^{2\times\textit{B}}$, where \textit{B} is the number of body parts with skin being learned in the experiment.
%\mh{WE CAN ADD THE 2xB BACK IF WE  USE THAT IN THE RESULTS -- TORSO AND HEAD CONCURRENTLY.}\fg{good point}

\subsection{Forward and inverse models}
\label{sec:forward_and_inverse_models}

Our focus is on inverse models: learning how to reach for particular locations on the skin ($\sim$ inverse kinematics).
From the models available in Explauto, the \textit{nearest neighbor} (NN) solution is the one used in our exploration strategies. Given an observation \textit{x}, the inverse model will return the motor command \textit{q} that corresponds to the observation stored in the database that is closest to \textit{x}. We do not employ forward models, with the exception of direct optimization on goal babbling (Section~\ref{sec:exploration strategies}) which uses them in a local optimization step.
%\mh{LIKE THIS WILL DO?} \fg{looks good}

%In the case of the forward model, given a motor command \textit{q}, the NN forward model will scan the database and return the observation \textit{x} corresponding to the motor command closest to \textit{q}.
%\mh{Are we using the forward model at all?} \fg{Indirectly when using the direct optimization on goal babbling}
%In our experiments, a forward model is not learned, apart for one of the method using temporary forward models.

%\mh{Similar to Action and observation spaces, list here the models that were used.}
%\fg{We only really used inverse model with NN, except when using direct optimization, which uses temporary forward models with LWLR for each goal. Do we need to add the definition of LWLR because of that?}

\subsection{Exploration strategies}
\label{sec:exploration strategies}

The exploration strategies from Explauto~\cite{Moulin2014} are:
\paragraph{Random Motor Babbling (RMB)} a motor configuration $\textit{q}\in\textit{Q}$ is sampled uniformly from the action space, and then executed, generating an observation $\textit{x}\in\textit{X}$.
\paragraph{Random Goal Babbling (RGB)} a goal $\textit{x}\in\textit{X}$ is sampled uniformly from the observation space and the inverse model is used to find an action $\textit{q}\in\textit{Q}$ best matching the goal, with added exploration noise.
\paragraph{Discretized Goal Babbling (DGB)} The interest space---the observation space in this case---is discretized into $\textit{c} = \textit{m} \times \textit{n}$ cells (or regions). We use two variants, 15x15 and 32x32 cells. Goal generation starts by selecting one of the cells with a probability proportional to the current state of an \textit{interest value} \textit{I} that each cell possesses. Then, a goal \textit{$x_c$} within that cell is uniformly generated. The robot attempts to reach for \textit{$x_c$} using inverse model prediction, resulting in the real observation \textit{$x$}.
\textit{I} is computed as the absolute value of the derivative of \textit{competence} \textit{C}. The interest value is high when competence rapidly increases or declines. \textit{C} is computed as follows: 

\begin{equation}
    \label{eq:1}
    \begin{array}{l}
    C \equiv d = ||x_c - x|| \\
    I =  |\frac{dC}{dt}|
  \end{array}
\end{equation}

DGB requires a bootstrapping phase to generate initial touches. This phase is counted towards the 1000 iteration limit. RMB with constrained joints range is used until 10 touches are observed. This is usually reached in 30 to 50 iterations.
\paragraph{Goal Babbling with direct optimization (DO)} Direct Optimization is an added layer on top of Goal Babbling strategies. For each generated goal and based on the inverse model prediction, a temporary forward model using Locally Weighted Linear Regression is created and optimized using Covariance Matrix Adaptation: Evolutionary Strategy for a set number of trials \textit{k} (10 in our experiments). The motor command with the observation closest to the goal is used to improve the main model, replacing the prediction.

\subsection{Learning and testing models}
\label{sec:learning_and_testing}

In each experiment, the robot uses a given exploration strategy and generates motor commands for 1000 steps. Unlike in standard cases in which every iteration of active exploration results in reaching a point in the observation space and allows for calculating an error (target vs. actual outcome---like in the case of learning to reach), in our case, the movement does not always result in contacting the skin. In that case, the step is counted toward the maximum number of iterations but does not contribute to learning the inverse model.

%, after which learning stops. Because the outcome of an action also depends on the initial joint position---the end posture should be the same, but collisions and touches could occur differently---we reset the position of the robot to a \textit{home posture}, avoiding wrong observations, as suggested in \cite{Rolf2012} p. 51. Because we want touch as the driver of body learning, only generated configurations that end up in at least one taxel being activated are kept for learning the inverse kinematics.

\begin{figure}[!ht]
\centering
\begin{subfigure}{.24\textwidth}
  \centering \includegraphics[scale=1, width=1\linewidth]{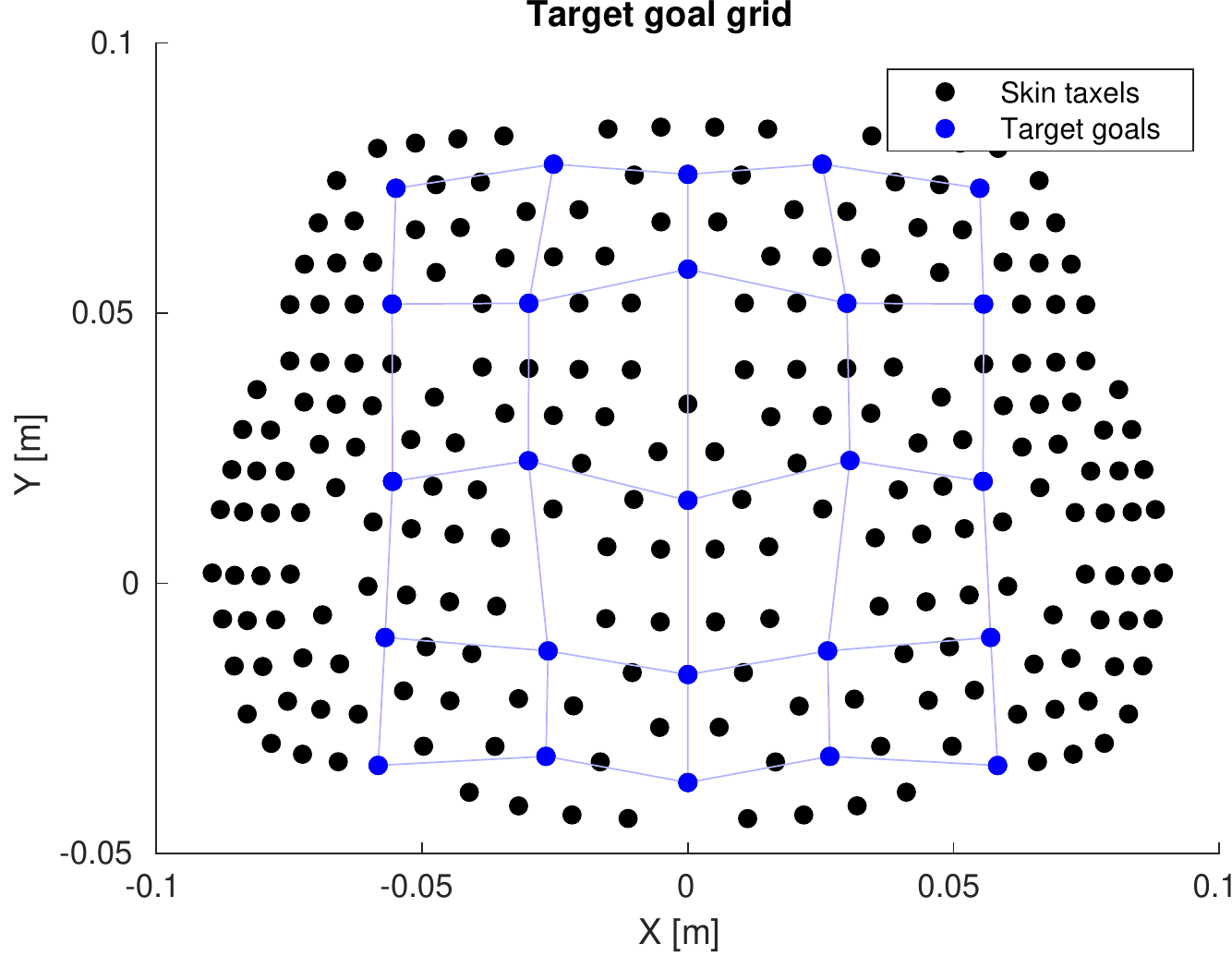}
\end{subfigure}%
\begin{subfigure}{.24\textwidth}
  \centering \includegraphics[scale=1, width=1\linewidth]{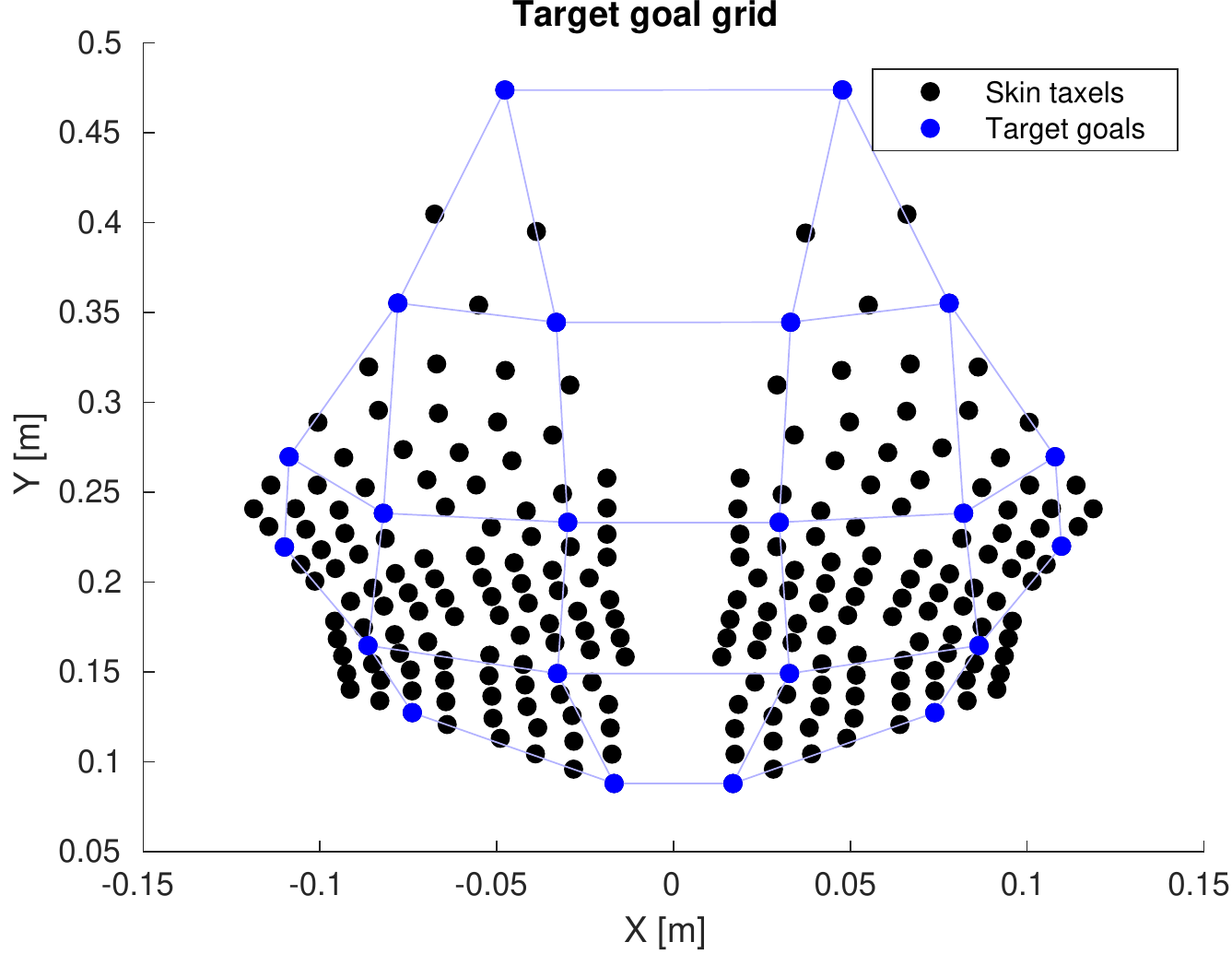}
\end{subfigure}
\caption{Testing set for high-resolution skin -- torso (left) -- head (right). All skin taxels in black; testing set grid in violet.}
%\caption{Testing set for high-resolution skin -- torso. All skin taxels in black; testing set grid in violet.}
\label{fig:test-grids}
\end{figure}

In addition, there is an external evaluation procedure that allows us to monitor the learning progress from the outside. Every 100 iterations, we present a testing set of taxels which the robot is asked to reach using the current inverse model.  
%The model is evaluated every 100 steps by trying to reach a set of taxels.
For the low-resolution skin, all taxels are included in the test set because of their already small number. For the high-resolution skin, a subset of taxels chosen to represent a grid-like pattern is tested (Fig.~\ref{fig:test-grids}).
Also here, if no taxel is contacted, no error can be measured. 
%Actions that did not generate a touch feedback, even if it touches the plastic casing close to the targeted taxel, mean the robot was not able to reach for the target, and no distance error is calculated, since there is no taxel to rely on for the calculation. As such, the mean reaching error (MRE) showed in the results (Figs.~\ref{fig:results-lowrestorso},~\ref{fig:results-lowreshead} and~\ref{fig:results-highrestorso}) only takes into account the cases when a touch has been generated and distance could be calculated between the reached and the goal taxels.

\section{Experiments and Results}
\label{sec:results}
We present results of a series of experiments. Some experiments are briefly illustrated also in the video at \url{https://youtu.be/Zb87uTFnQZE}. The right hand will be the robot's effector and will reach either for the torso skin (Section~\ref{sec:right_hand_torso}) or for the head skin (Section~\ref{sec:right_hand_head}). In the latter case, the action space will be larger as two neck joints are available in addition to the five arm joints. We use both versions of the skin: low-resolution and high-resolution (Fig.~\ref{fig:naogazebo}). The low-res. skin has the practical advantage that experiments are considerably faster (high-res. skin emulation is computationally expensive). However, the high-res. skin more closely models the real robot's shape. Comparing these two cases is interesting in itself: the observation space is larger (more taxels), but due to their higher density, it is ``smoother'' since errors can be more accurately measured.

%The result section is separated per body part involved, and per resolution of the skin, as the observation space are different and the comparison of results more appropriate to be done within the same space.

We will illustrate the results in the following ways: (i) Mean Reaching Error (MRE) after every 100 iterations (e.g., Fig.~\ref{fig:results-lowrestorso}, left), (ii) number of touches generated after every 100 iterations (e.g., Fig.~\ref{fig:results-lowrestorso}, right), (iii) projection of the generated goals with details about the reaching error for each test taxel (after 1000 iterations; e.g., Fig.~\ref{fig:proj-lowres-torso}). The results are averaged over ten trials for each exploration strategy.
For projections, the observation space is presented from the point of view of an observer looking at the robot---like in Fig.~\ref{fig:naogazebo}. DO methods were not run on high-res. skin.
%Therefore, the right side from the robot's PoV is on the left side on the shown projections, as it is seen from an observer's PoV.
Note that reaching errors are only available when a taxel (target or other) was reached by the movement.
%The results include the comparison of the mean reaching error (MRE), the number of touches generated by the methods, and for goal babbling methods, a projection of the generated goals with details about the reaching error for each taxel, if the inverse model generated a touch feedback when trying to reach them. These projections only show the last test phase results, after all learning steps.

\subsection{Right hand reaching for torso}
\label{sec:right_hand_torso}

\begin{figure}[!ht]
\centering
\begin{subfigure}{0.24\textwidth}
  \centering \includegraphics[scale=1, width=1\linewidth]{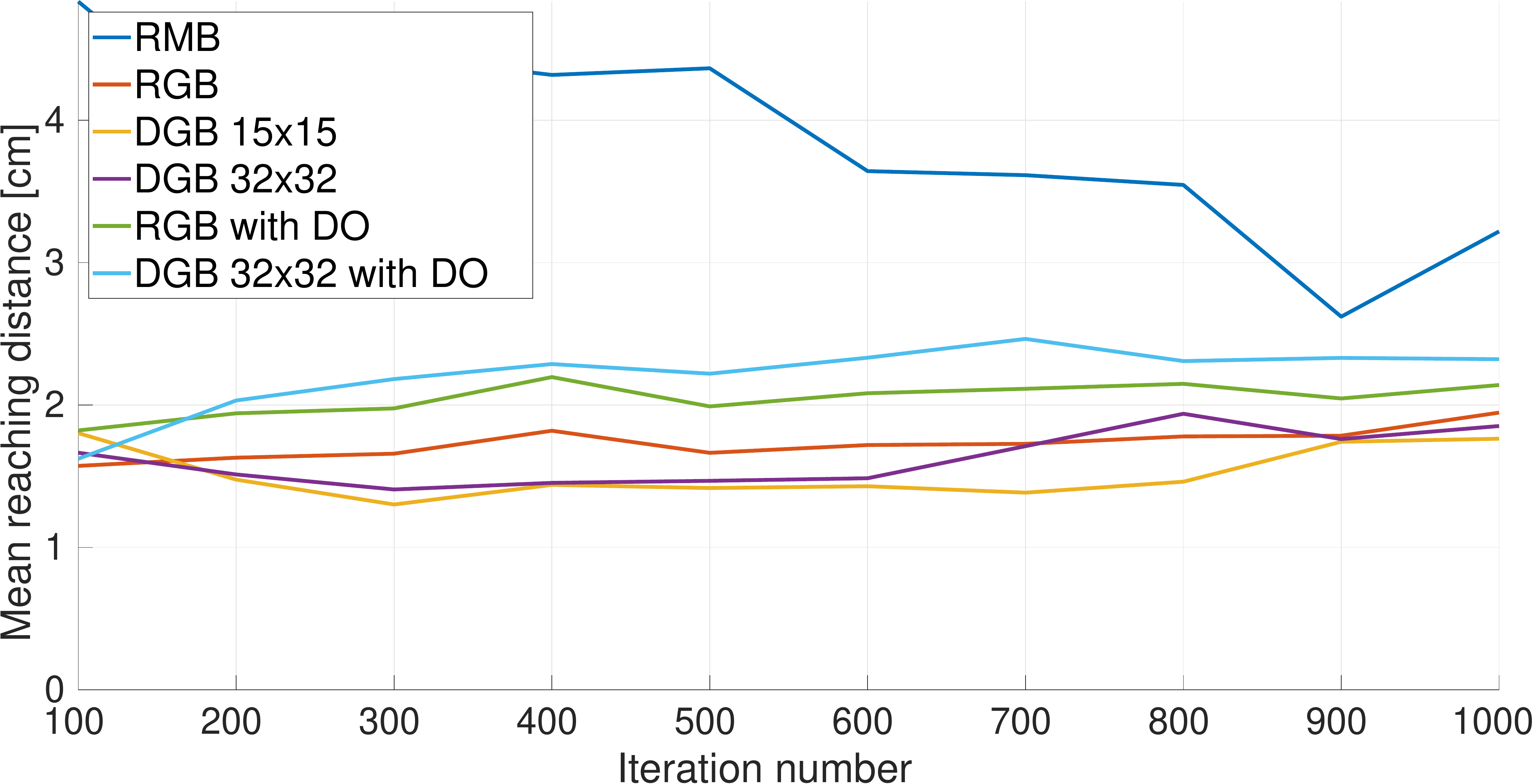}
\end{subfigure}%
\begin{subfigure}{0.24\textwidth}
  \centering \includegraphics[scale=1, width=1\linewidth]{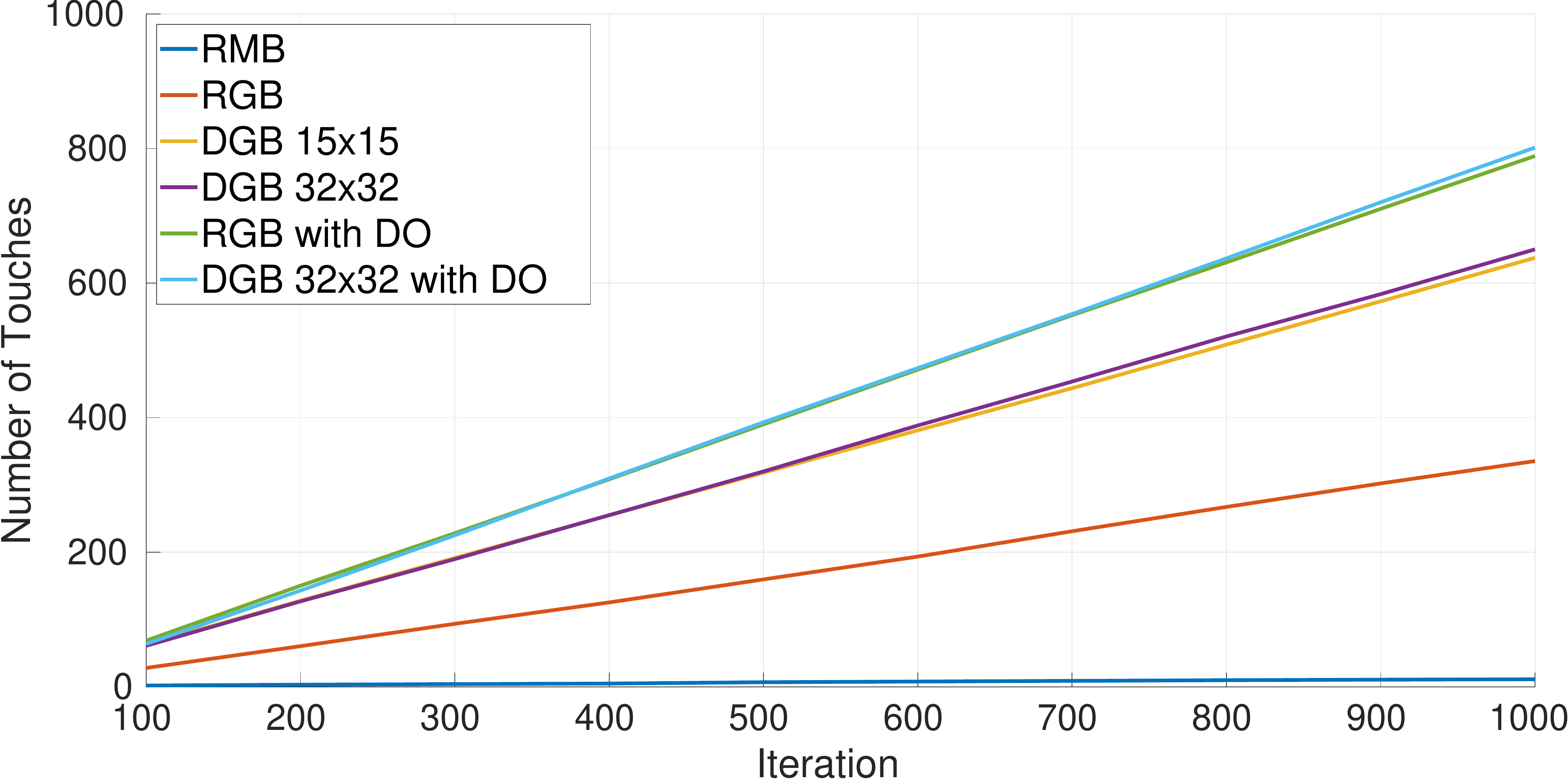}
\end{subfigure}
\caption{Right hand reaching for torso, low-res. skin -- comparison of exploration strategies (descriptions in Section~\ref{sec:exploration strategies}). (Left) MRE. (Right) Number of touches generated.}
\label{fig:results-lowrestorso}
\end{figure}

\paragraph{Low-resolution skin}
When reaching for the torso, RMB and DGB with DO show the highest MRE (Fig.~\ref{fig:results-lowrestorso} left), while the other methods have similar performance. The MRE does not decrease over time, which is somewhat surprising, but may be caused by the fact that, initially, during performance evaluation, the skin is not reached at all and hence no error is measured for some taxels. Later, taxels other than target taxels may be reached more frequently, thus contributing to MRE. Both DGB strategies without DO have similar number of touches, like the two methods with DO, while RMB generates almost none (Fig.~\ref{fig:results-lowrestorso} right). Despite having a higher number of touches, DGB 32x32 with DO has higher MRE throughout the experiments than any other method except RMB.

\begin{figure}[!ht]
\centering
\begin{subfigure}{.24\textwidth}
  \centering \includegraphics[scale=1, width=1\linewidth]{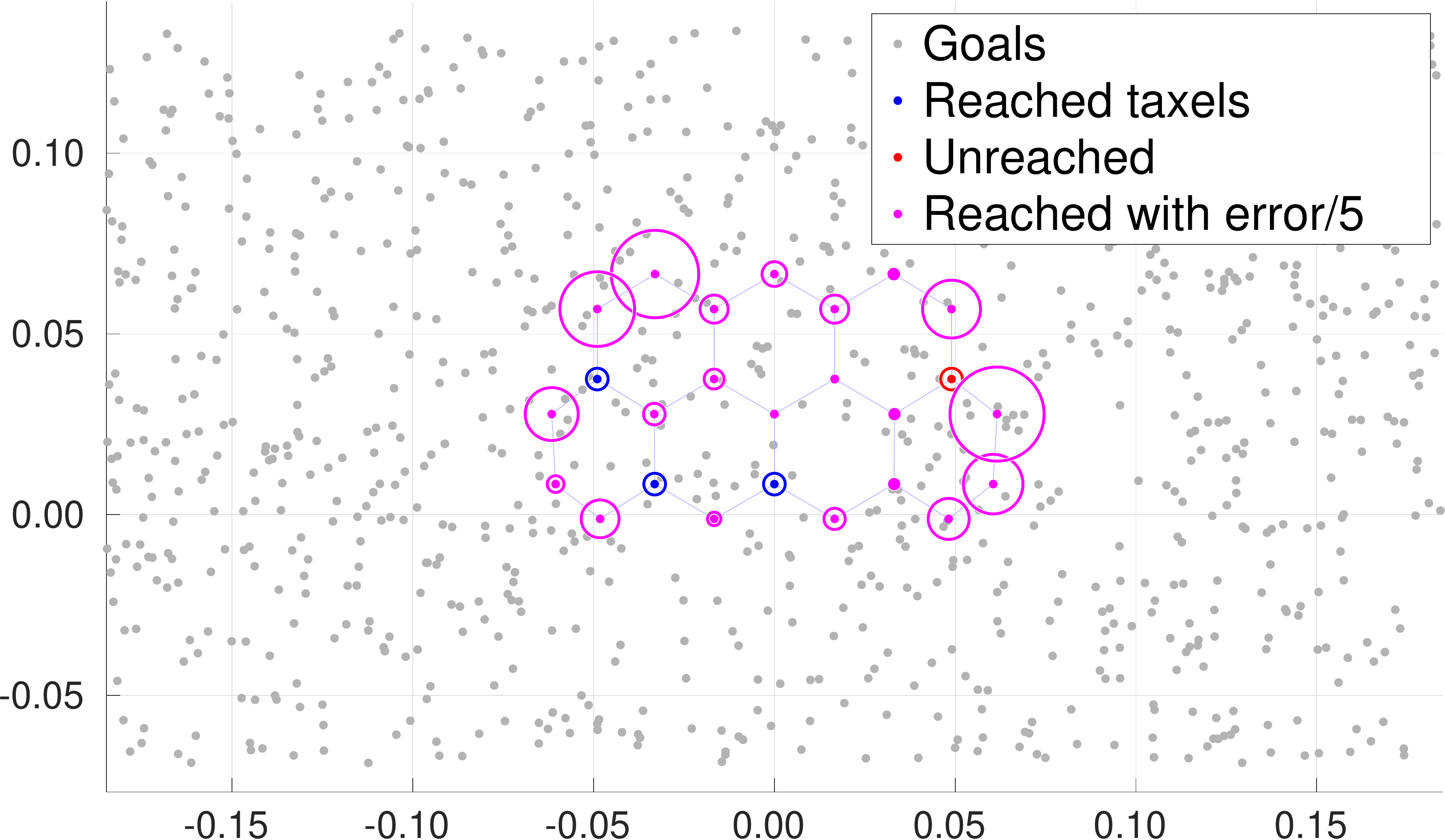}
  (a) RGB
\end{subfigure}%
\begin{subfigure}{.24\textwidth}
  \centering \includegraphics[scale=1, width=1\linewidth]{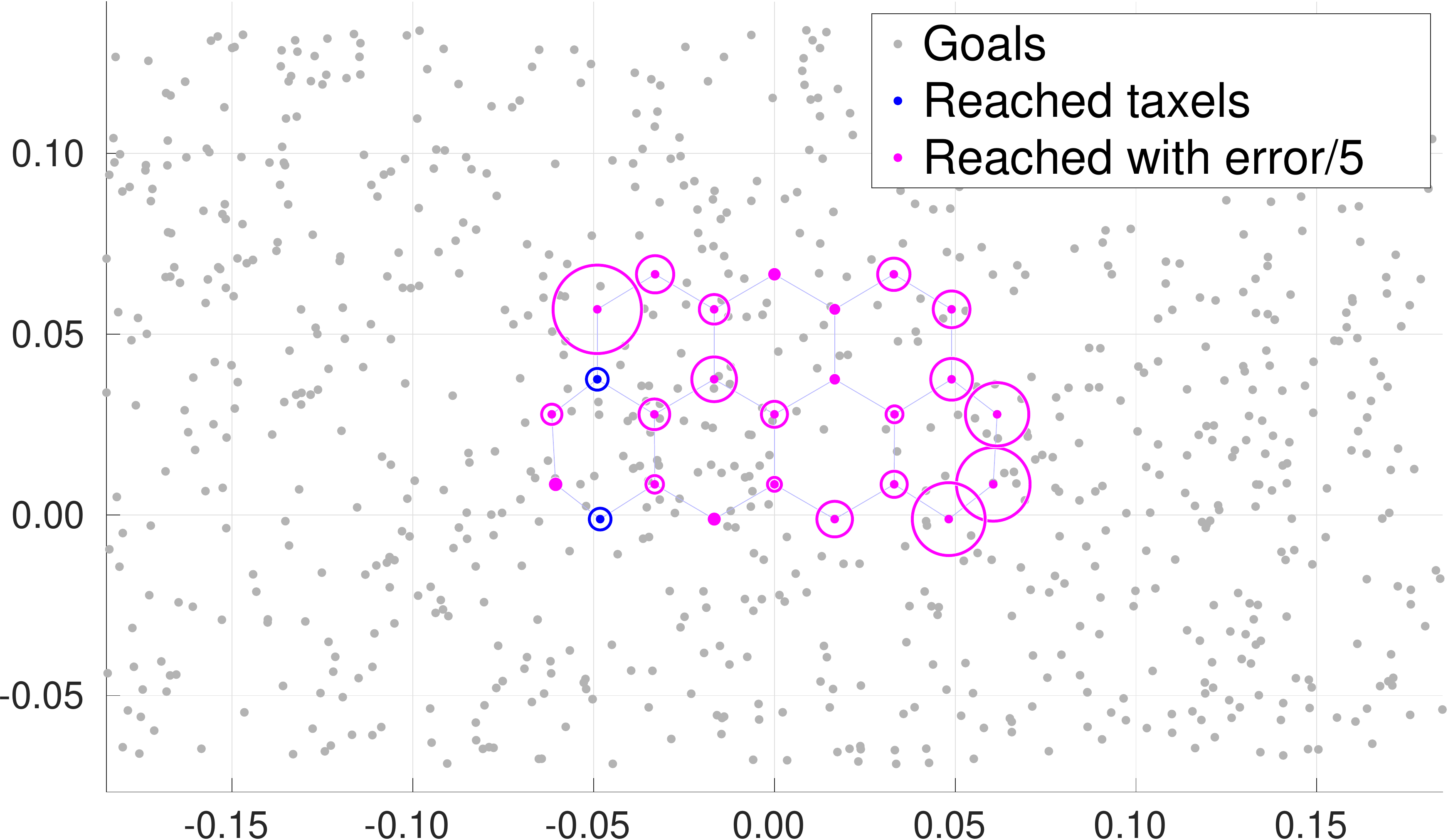}
  (b) RGB with DO
\end{subfigure}%

\begin{subfigure}{.24\textwidth}
  \centering \includegraphics[scale=1, width=1\linewidth]{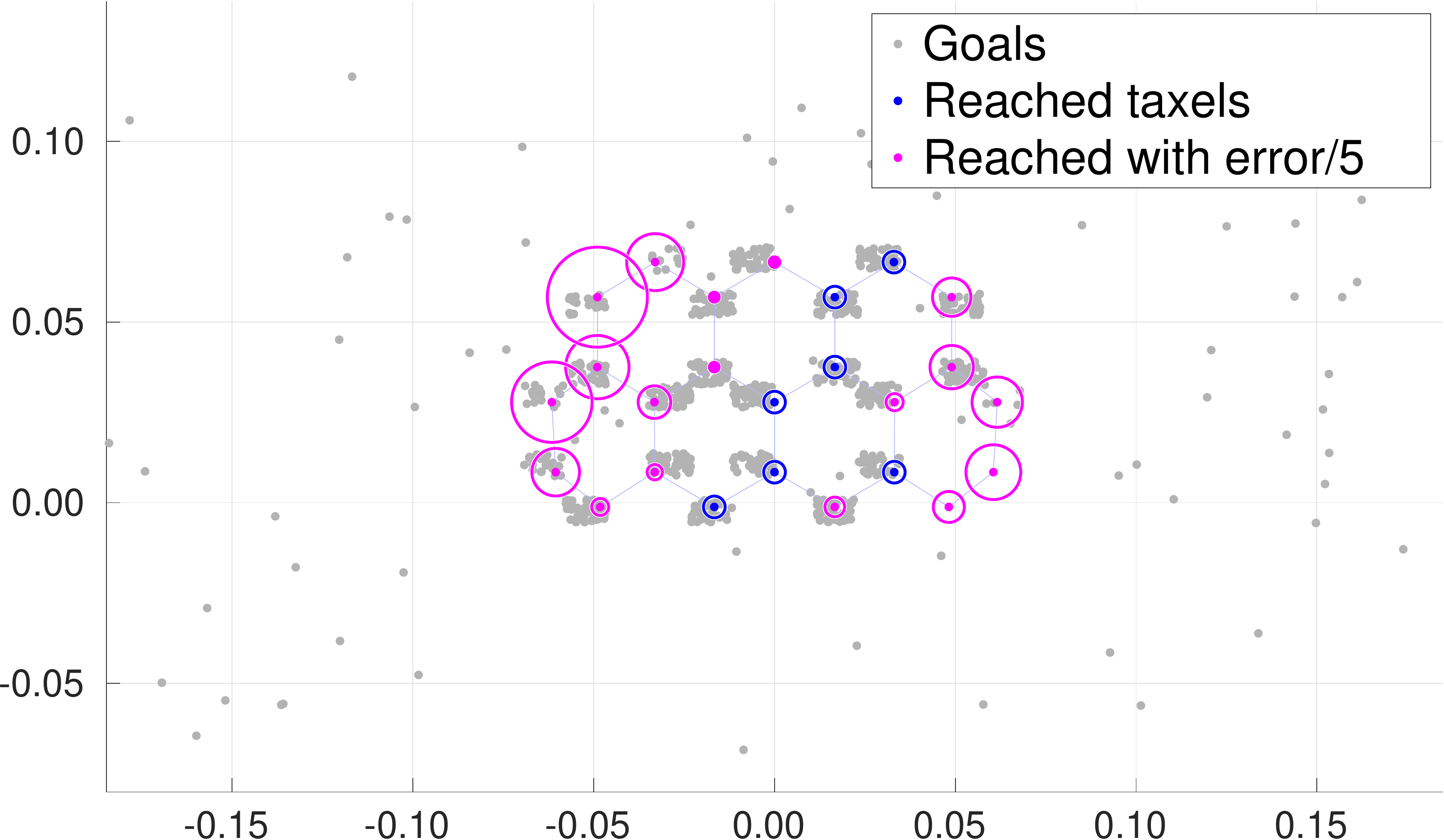}
  (c) DGB 32x32
\end{subfigure}
\begin{subfigure}{.24\textwidth}
  \centering \includegraphics[scale=1, width=1\linewidth]{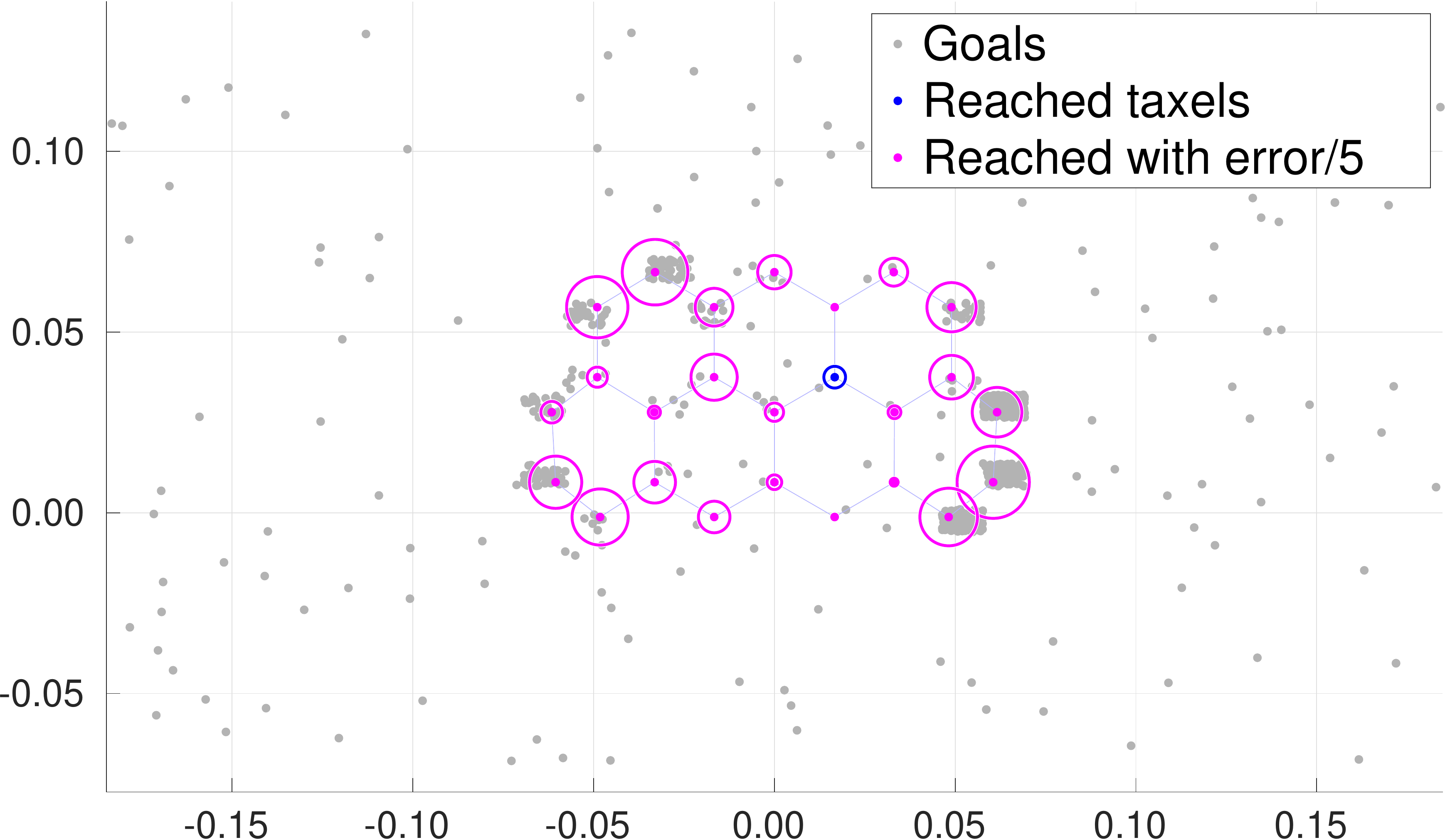}
  (d) DGB 32x32 with DO
\end{subfigure}%

\caption{Right hand reaching for torso, low-res. -- observation space. \textit{Goals} generated during exploration process (grey); testing after all learning iterations: \textit{Reached taxels} with no error (blue); \textit{Reached with error / 5} -- taxels reached (magenta dots) with error magnitude divided by 5 (magenta circles); \textit{Unreached} taxels -- no taxel reached during reaching attempt (red).}
\label{fig:proj-lowres-torso}
\end{figure}

To study how the methods covered the observation space and how accurate is reaching to different parts of the skin after learning, we use the observation space plots -- Fig.~\ref{fig:proj-lowres-torso}. The panels also show the reaching performance to individual taxels after learning finished. Some taxels can be reached with no error over all trials (blue); for most taxels, different taxels than asked are reached, giving rise to errors (magenta circles around target taxel). Finally, for some reaching targets, no contact with skin is detected and no error can be measured (red). To avoid mis-representing errors for taxels that are often ``unreached'' (i.e., no contact with skin occurs and error cannot be measured), only taxels that have been reached (perfectly or with error) in 60\% or more trials have their reaching error calculated.  The magenta circles displayed in their full extent are hard to interpret in some projections; therefore, circles are rendered with a radius of one fifth of the reaching error.
%To improve the readability of the projections, the reaching error was divided by 5. The full error for RGB can be seen in Fig.~\ref{fig:proj-lowres-torso} a), while the reduced error for the same experiment is shown in (Fig.~\ref{fig:proj-lowres-torso}) b).
The projection of all goals generated is uniformly sampled from the pool of goals generated by all ten trials.

For the case of right hand reaching for torso, low-res. skin, Fig.~\ref{fig:proj-lowres-torso} shows that Random Goal Babbling (panel a) has a uniform distribution of goals.
DGB methods focus the exploration best, despite some missed spots.
%generates much more meaningful goals, closer to taxels, compared to RGB. 
However, better goal generation does not automatically mean that the robot was able to successfully reach for them: it might have reached for a close taxel, or even for a space between taxels, leading to no touch feedback and no learning. We can observe this by looking at the projection of individual trials, or at Fig.~\ref{fig:proj-lowres-torso}, d), where a taxel on the bottom right with numerous close goals has an on average higher error than other taxels whose neighborhood was less frequently sampled (with goals). 
Simple RGB displays good results: It can on average correctly reach around half of the taxels, as we can see a lot of taxels with errors so small that only a dot can be seen and not the dot and the error circle separately. DGB displays better results, with more taxels being consistently reached over the trials, and other taxels reached with small errors (often because in one or two of the trials, it reached for the closest taxel instead of the target taxel).
Surprisingly, direct optimization did not improve the performance results. Over the course of the trials, it seems that only for RGB a particular taxel stayed mostly unreached. For both RGB and DGB, the taxels with the highest reaching errors are on the sides of the skin.

\paragraph{High-resolution skin}
%High-resolution skin requires more computing power and time to run the simulation, and choices about what experiments to run had to be made. Direct Optimization methods were not run with the high-resolution skin, as they displayed mitigated results on the low-resolution skin and they multiply by a ten-factor (with our current settings) the running time of the learning phase. However, these methods might be more effective on high-resolution skin than on low-resolution, due to the higher density of taxels. \fg{This last sentence should probably not be here, but rather on the discussion or somewhere else}

\begin{figure}[!ht]
\centering
\begin{subfigure}{.24\textwidth}
  \centering \includegraphics[scale=1, width=1\linewidth]{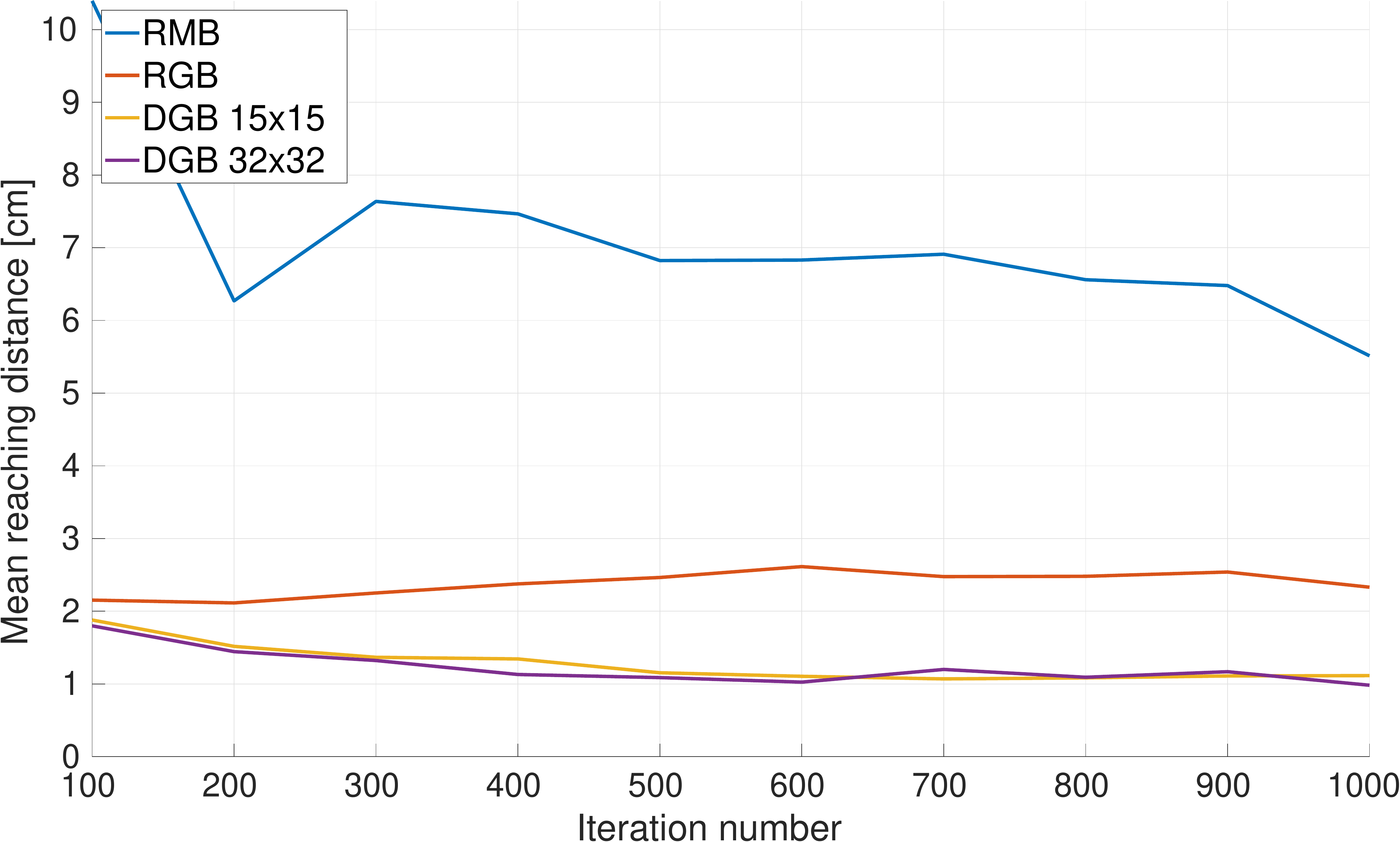}
\end{subfigure}%
\begin{subfigure}{.24\textwidth}
  \centering \includegraphics[scale=1, width=1\linewidth]{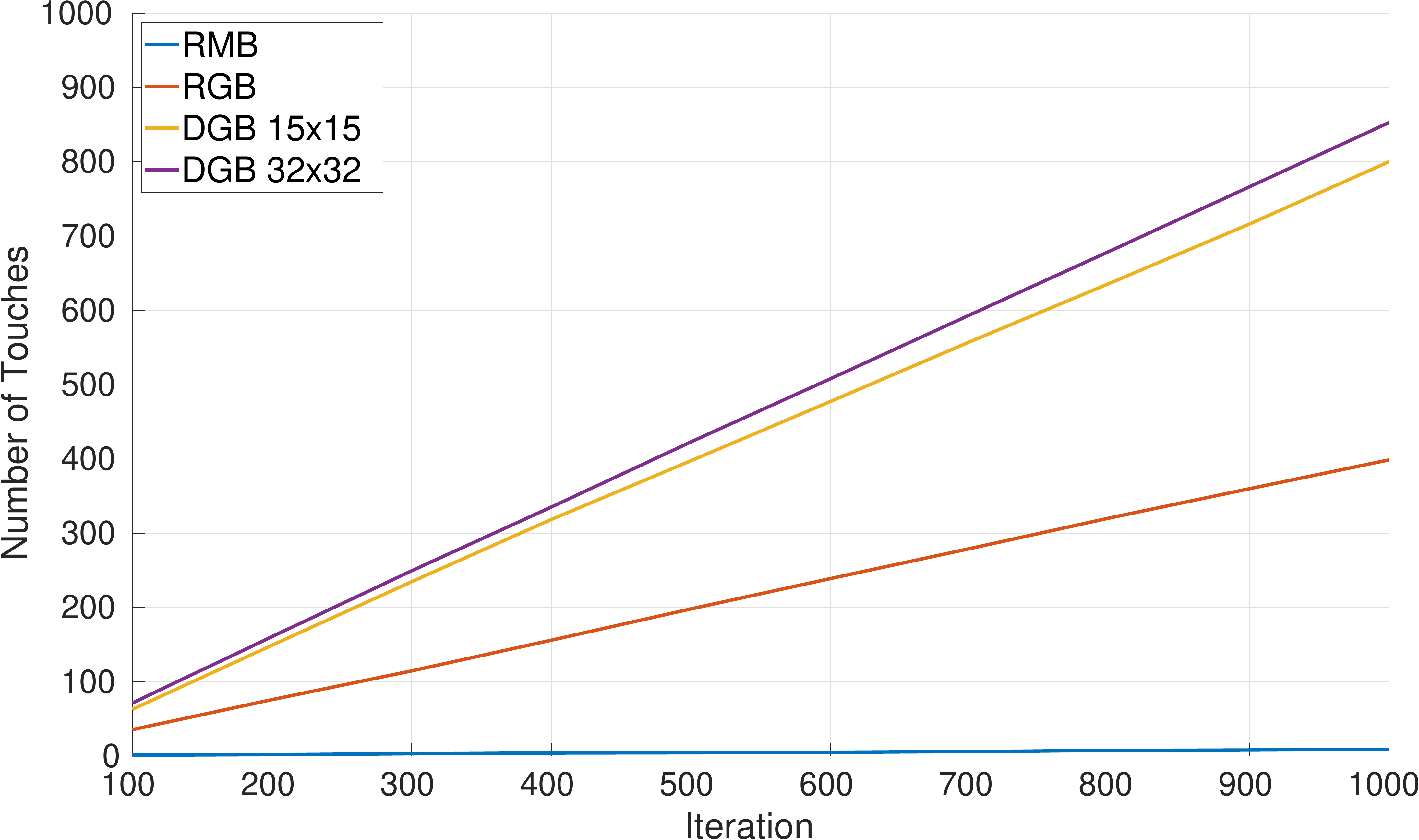}
\end{subfigure}
\caption{Right hand reaching for torso, high-res. skin. (Left) MRE. (Right) Number of touches generated.}
\label{fig:results-highrestorso}
\end{figure}

Compared to the low-res. skin, the MRE (Fig.~\ref{fig:results-highrestorso}, left) starts at the same levels, but shows a clear decrease of the error for DGB methods, with higher number of touches (Fig.~\ref{fig:results-highrestorso}, right). This may be due to the higher density of taxels: the distances between taxels are lower. There is also a small effect on the number of touches, with DGB 32x32 generating more than DGB 15x15.

\begin{figure}[!ht]
\centering
%\begin{subfigure}{.24\textwidth}
%  \centering \includegraphics[scale=1, width=1\linewidth]{figures/Results/High_res/Torso/Goals/exps_average_results/MeanProjection_highres_torso_RMB.pdf}
%  (a) RMB
%\end{subfigure}%
\begin{subfigure}{.24\textwidth}
  \centering \includegraphics[scale=1, width=1\linewidth]{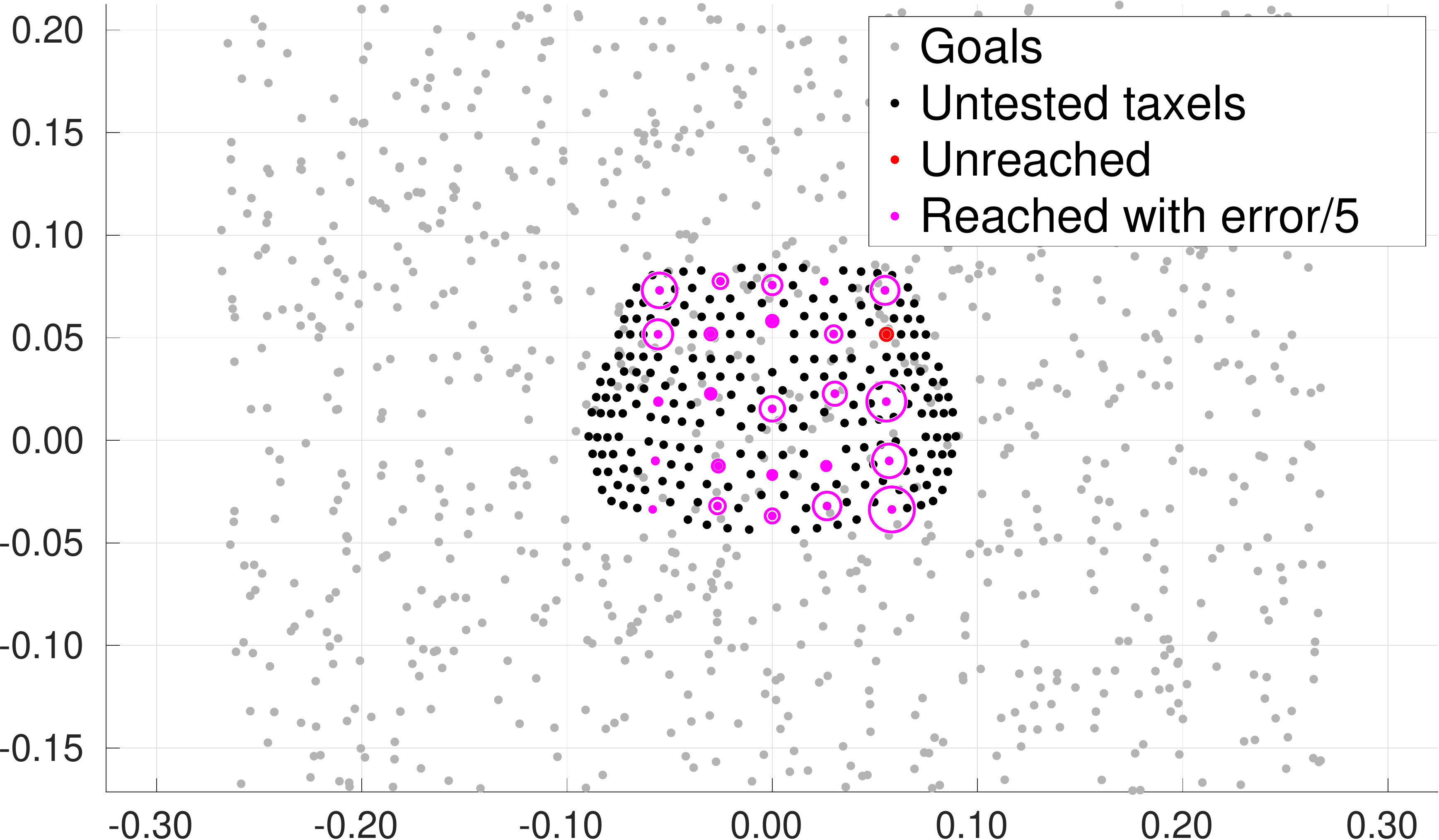}
  (a) RGB
\end{subfigure}%

\begin{subfigure}{.24\textwidth}
  \centering \includegraphics[scale=1, width=1\linewidth]{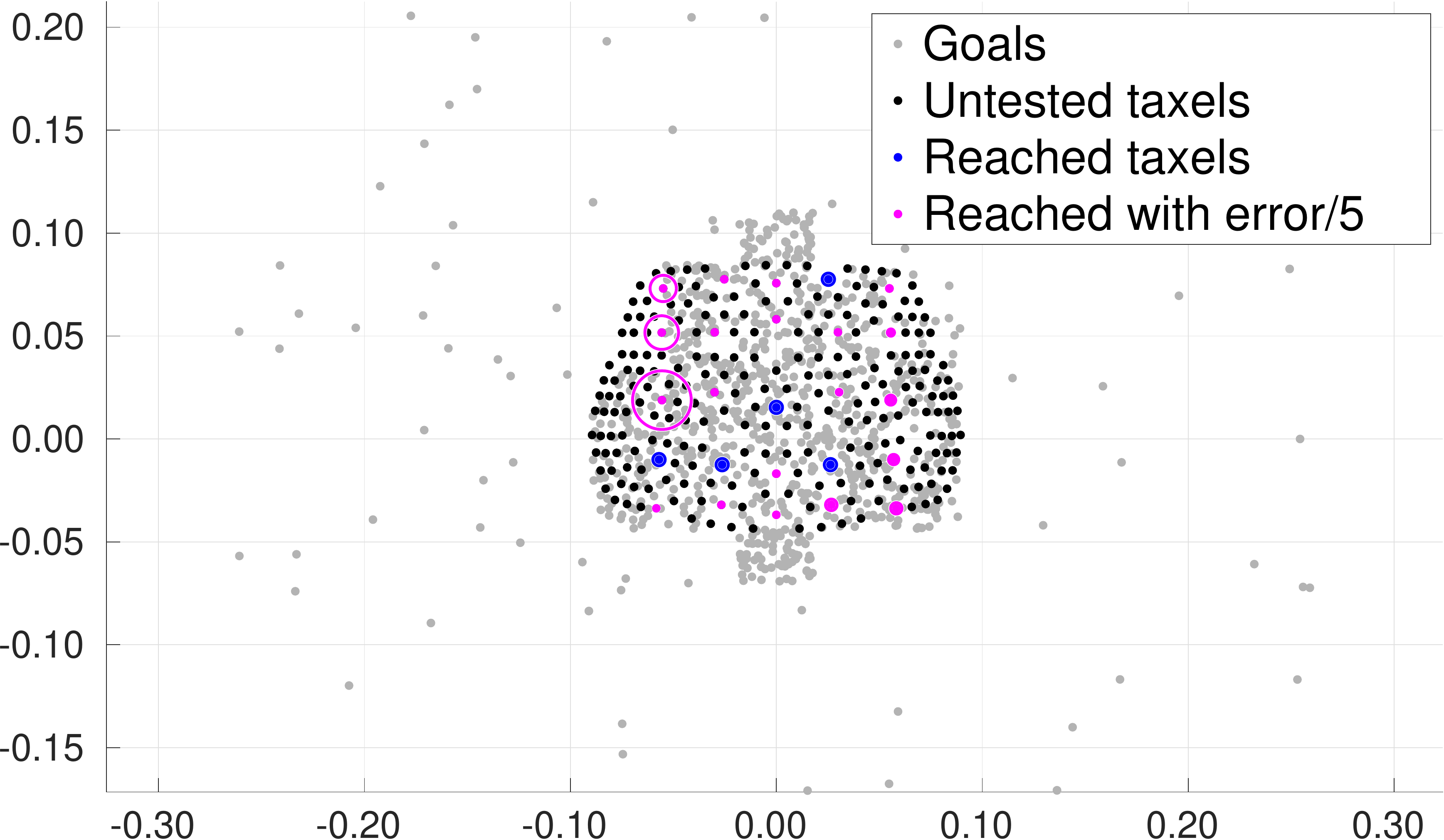}
  (b) DGB 15x15
\end{subfigure}
\begin{subfigure}{.24\textwidth}
  \centering \includegraphics[scale=1, width=1\linewidth]{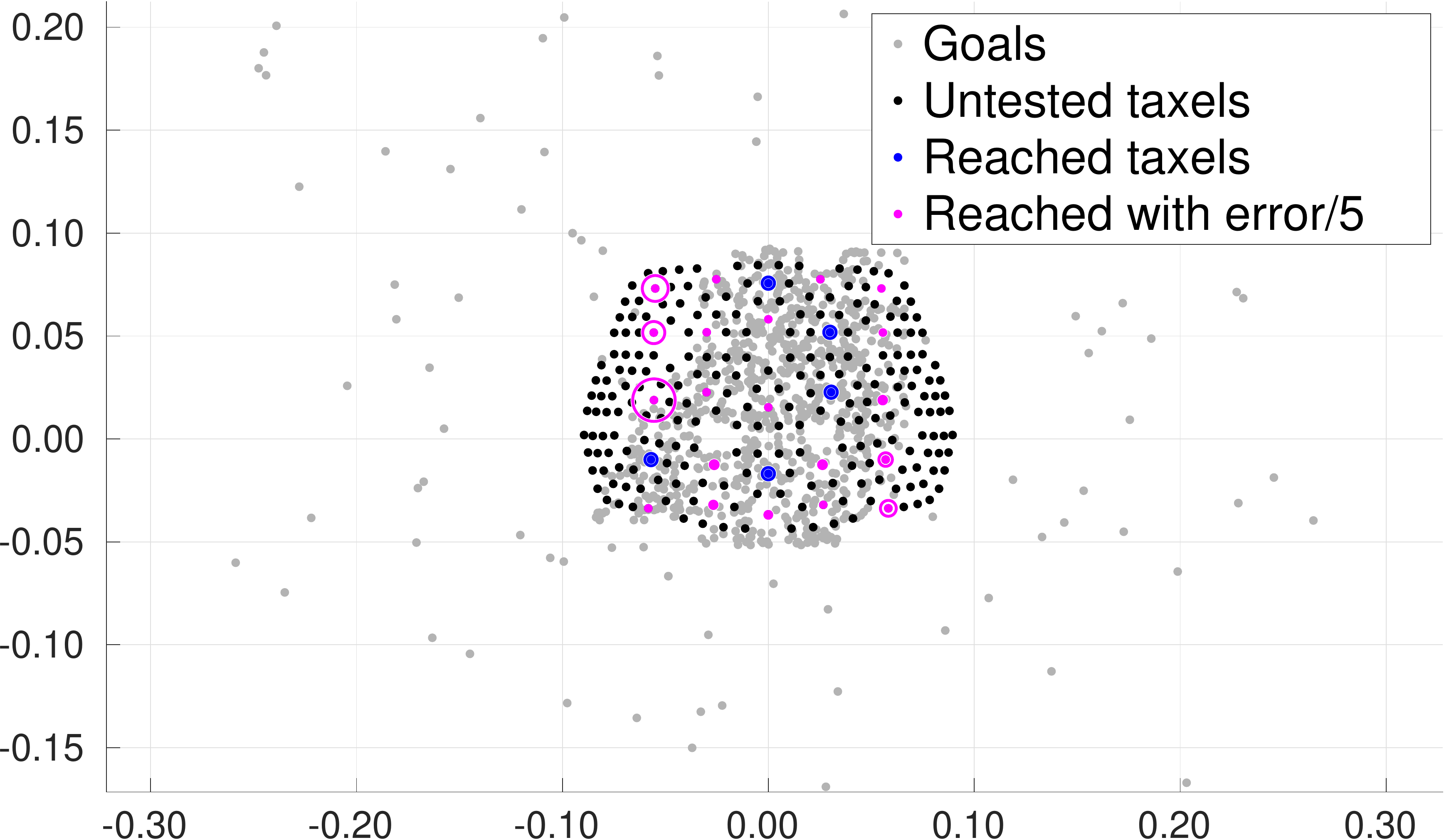}
  (c) DGB 32x32
\end{subfigure}

\caption{Right hand reaching for torso, high-res. skin. All skin taxels shown with black dots. See Fig.~\ref{fig:proj-lowres-torso} for details.}

\label{fig:proj-highres-torso}
\end{figure}

Fig.~\ref{fig:proj-highres-torso} visualizes the observation space. As explained in Section~\ref{sec:right_hand_torso}, at least 60\% trials contacting the skin---target or other taxel---are needed for average error to be calculated out of these trials. Random Motor Babbling (RMB) is not shown---no goals can be displayed as exploration proceeded in the motor space and reaching errors could not be measured for any of the taxels. DGB methods show again several consistently perfectly reached taxels or with low error. The goal generation from DGB 15x15 seems to be slightly more spread out over the skin than DGB 32x32. The taxels with the most errors are on the sides, like for the low-res skin, and the highest errors are on the same set of taxels.

\subsection{Right hand reaching for head}
\label{sec:right_hand_head}

\paragraph{Low-resolution skin}

\begin{figure}[!ht]
\centering
\begin{subfigure}{.24\textwidth}
  \centering \includegraphics[scale=1, width=1\linewidth]{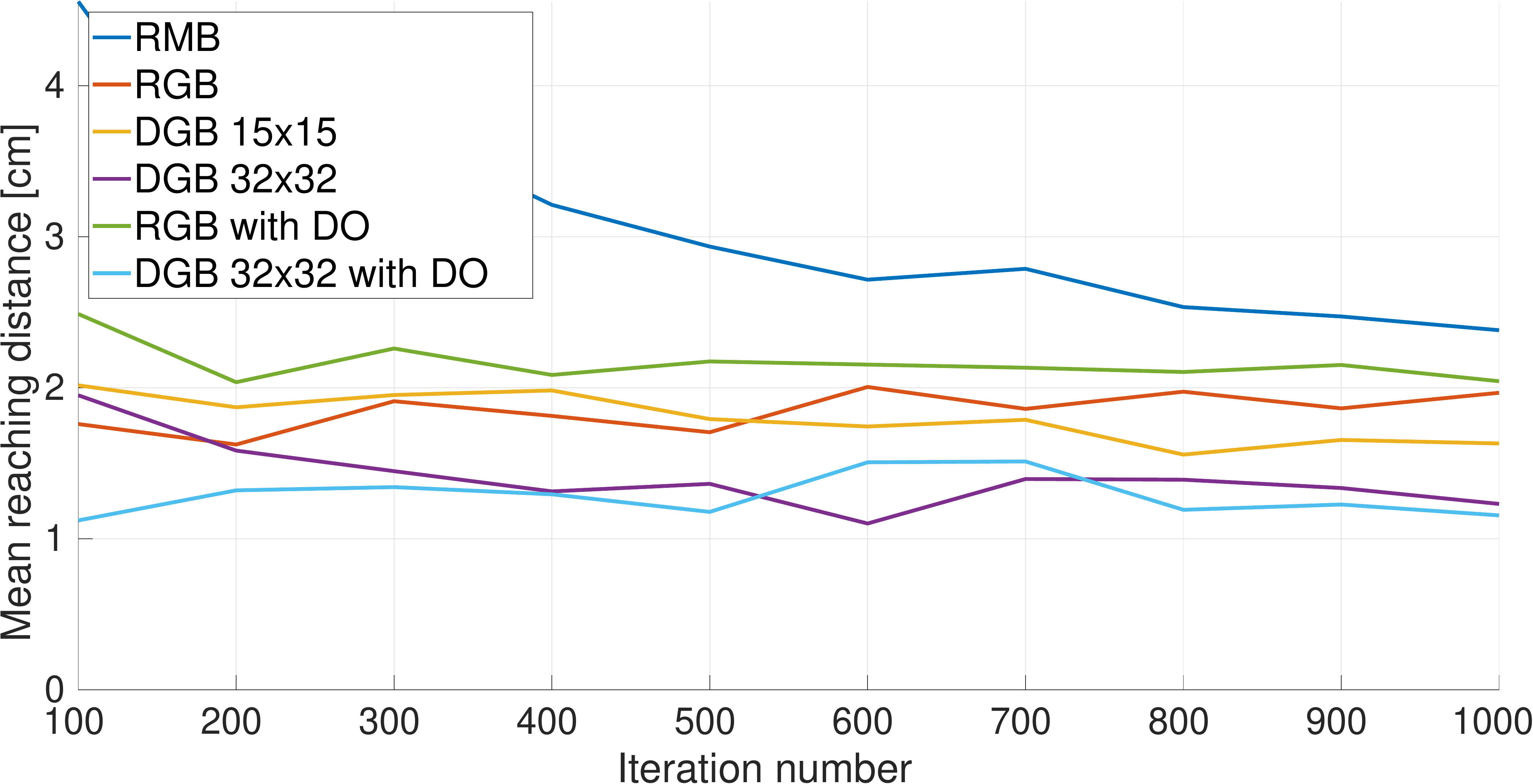}
\end{subfigure}%
\begin{subfigure}{.24\textwidth}
  \centering \includegraphics[scale=1, width=1\linewidth]{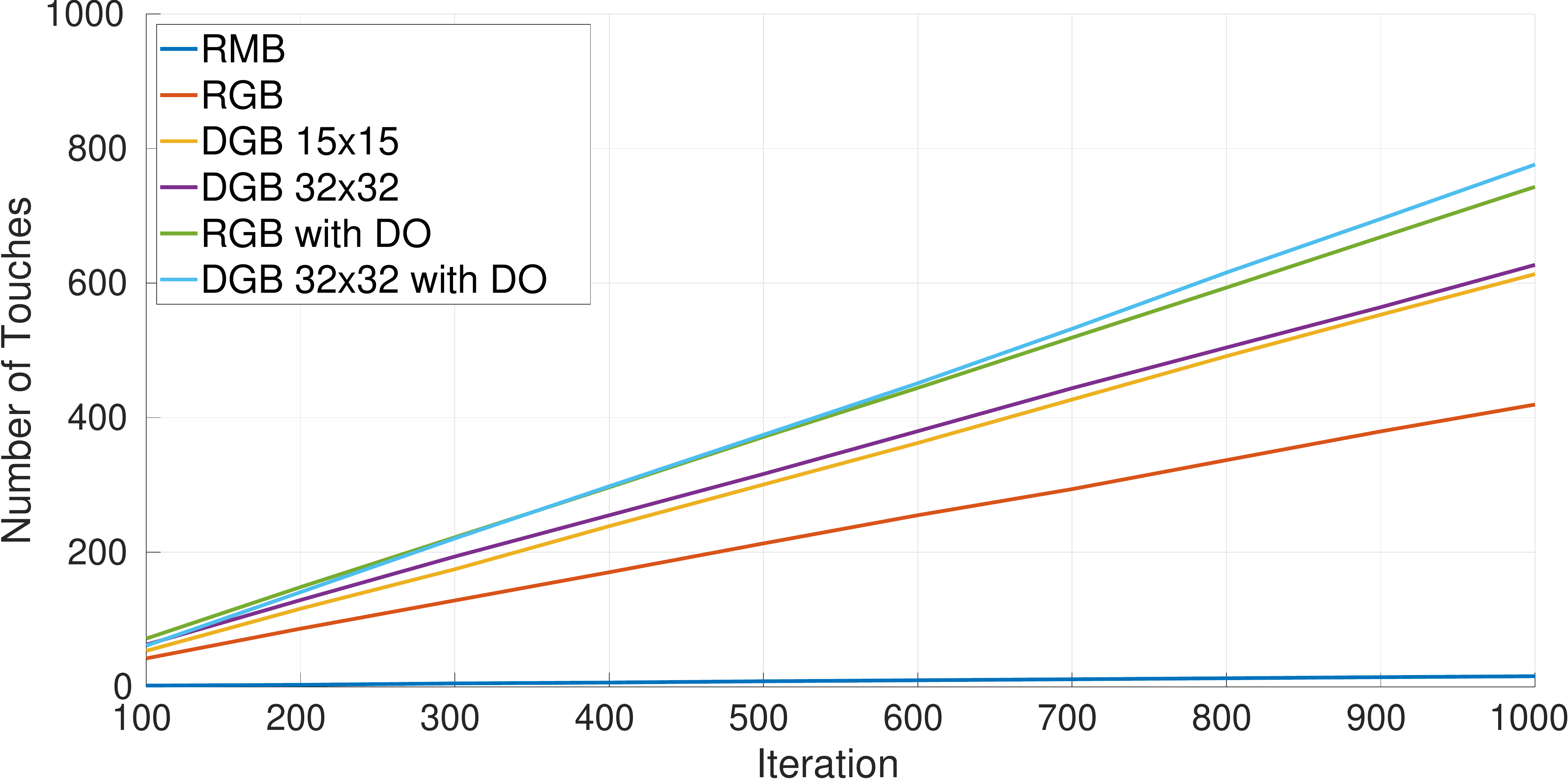}
\end{subfigure}
\caption{Right hand reaching for head, low-res. skin. (Left) MRE. (Right) Number of touches generated.}
\label{fig:results-lowreshead}
\end{figure}

Results on the low-res. head (Fig.~\ref{fig:results-lowreshead}) show similar MRE as well as number of touches as the low-res. torso results. However, DGB 32x32 with DO performs better and is one of the methods with the lowest MRE, along with its counterpart without DO.

\begin{figure}[!ht]
\centering
\begin{subfigure}{.24\textwidth}
  \centering \includegraphics[scale=1, width=1\linewidth]{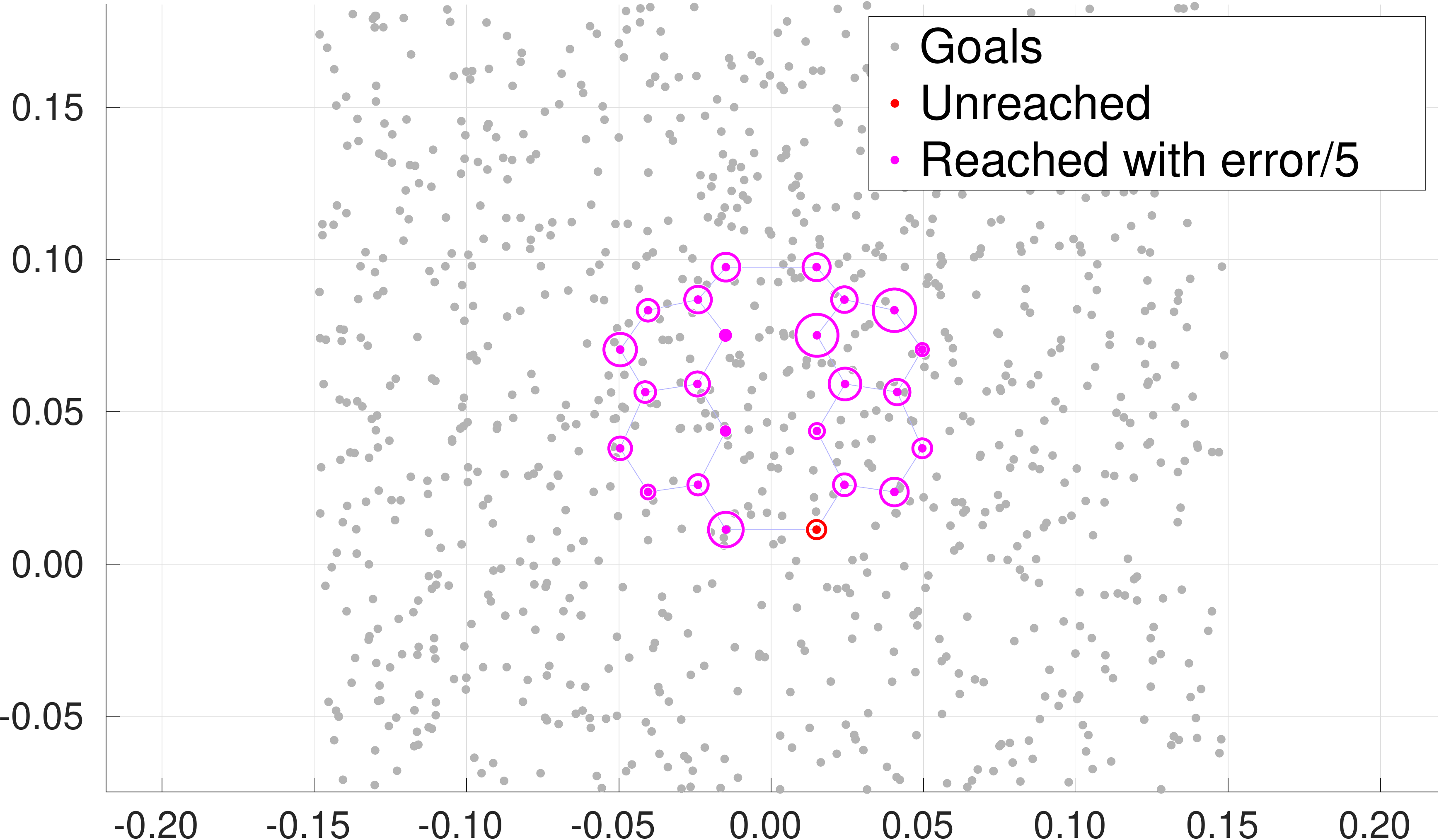}
  (a) RGB
\end{subfigure}%
%\begin{subfigure}{.24\textwidth}
%  \centering \includegraphics[scale=1, width=1\linewidth]{figures/Results/Low_Res/Head/Goals/exps_average_results/MeanProjection_lowres_head_RGBDO.pdf}
%  (b) RGB with DO
%\end{subfigure}
\begin{subfigure}{.24\textwidth}
  \centering \includegraphics[scale=1, width=1\linewidth]{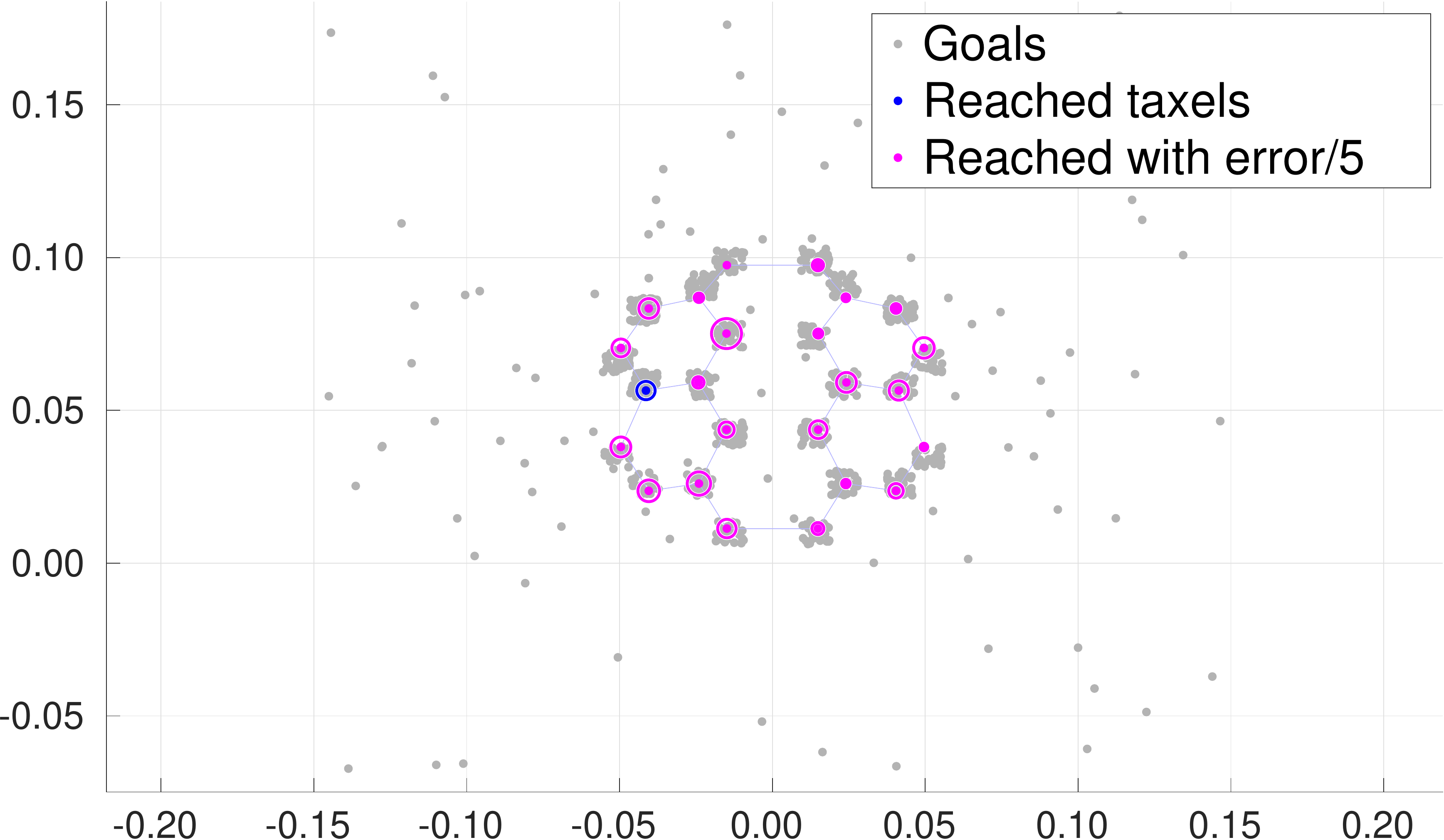}
  (b) DGB 32x32
\end{subfigure}%

%\begin{subfigure}{.24\textwidth}
%  \centering \includegraphics[scale=1, width=1\linewidth]{figures/Results/Low_Res/Head/Goals/exps_average_results/MeanProjection_lowres_head_DGB32DO.pdf}
%  (d) DGB 32x32 with DO
%\end{subfigure}
\begin{subfigure}{.24\textwidth}
  \centering \includegraphics[scale=1, width=1\linewidth]{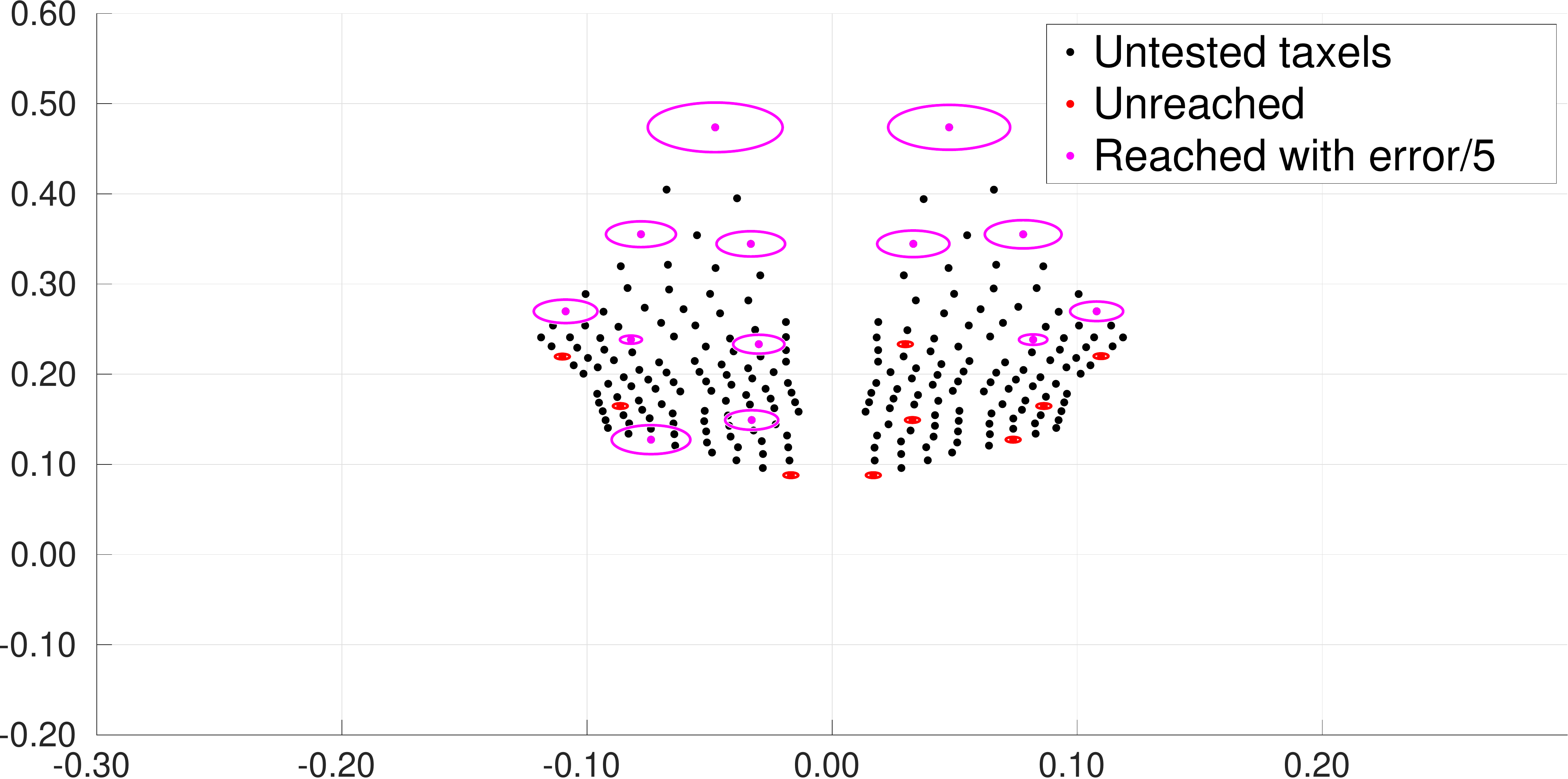}
  (c) RMB--high res.
\end{subfigure}
\begin{subfigure}{.24\textwidth}
  \centering \includegraphics[scale=1, width=1\linewidth]{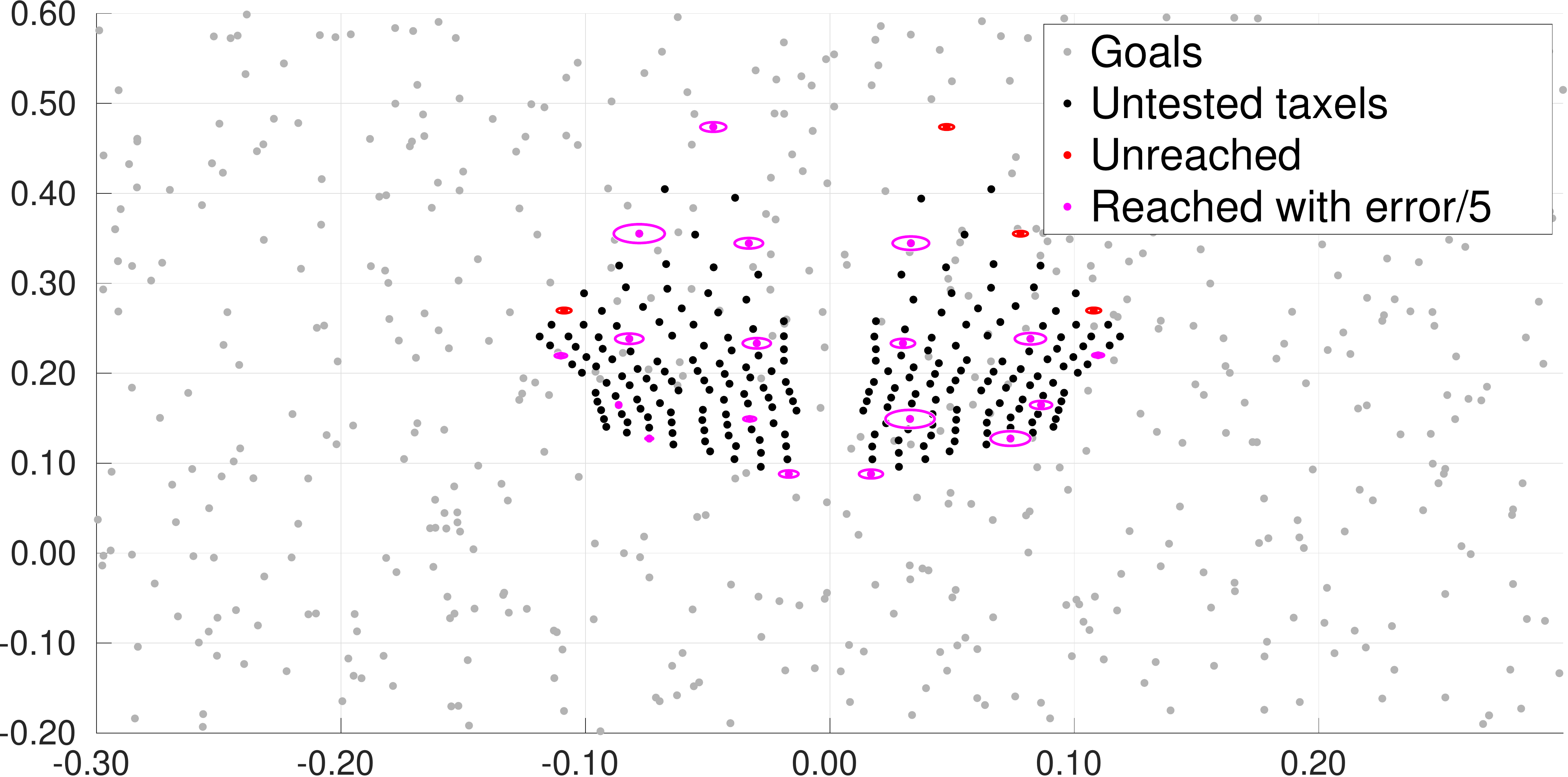}
  (d) RGB--high res.
\end{subfigure}%

\begin{subfigure}{.24\textwidth}
  \centering \includegraphics[scale=1, width=1\linewidth]{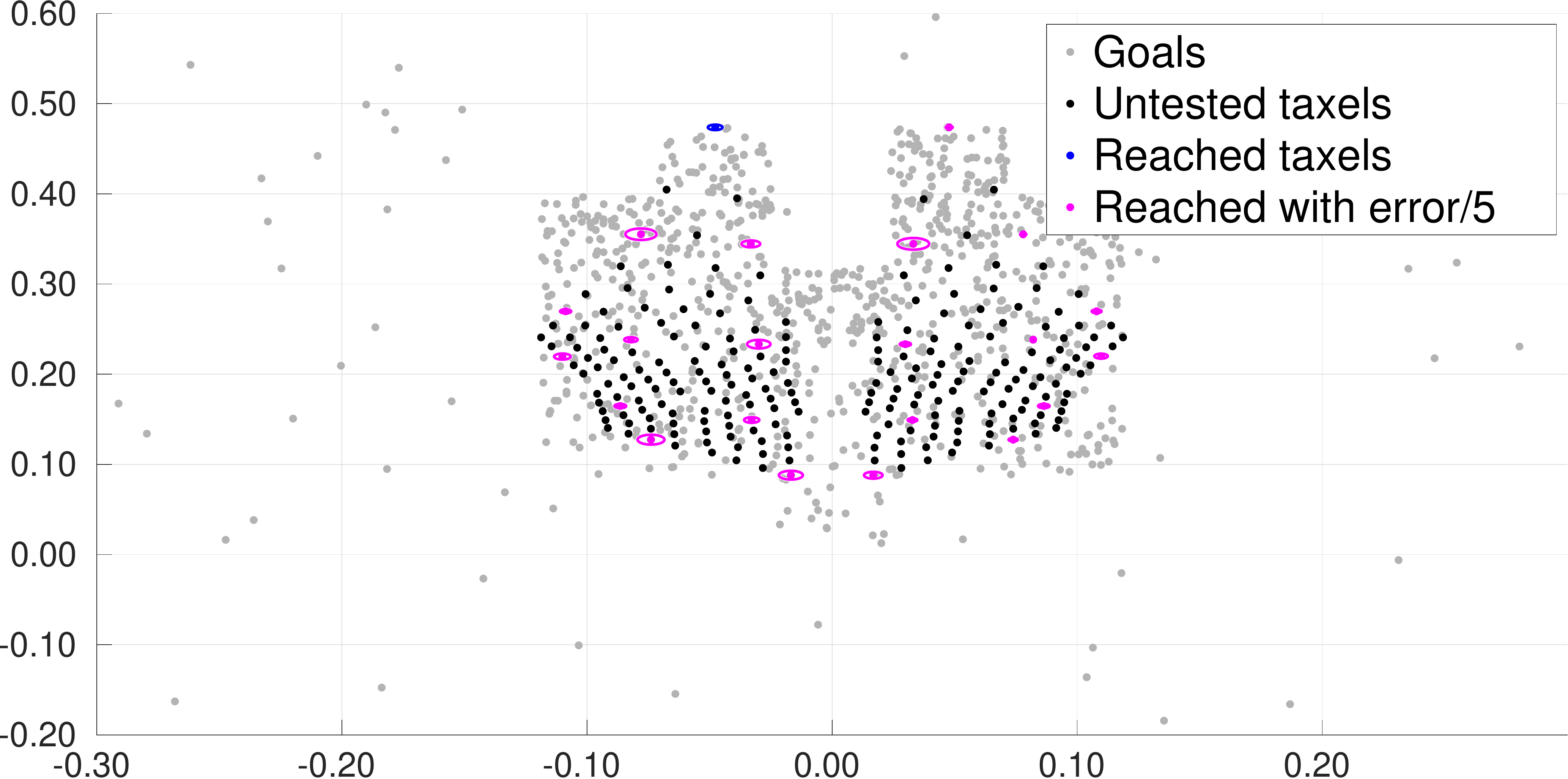}
  (e) DGB 15x15--high res.
\end{subfigure}%
\begin{subfigure}{.24\textwidth}
  \centering \includegraphics[scale=1, width=1\linewidth]{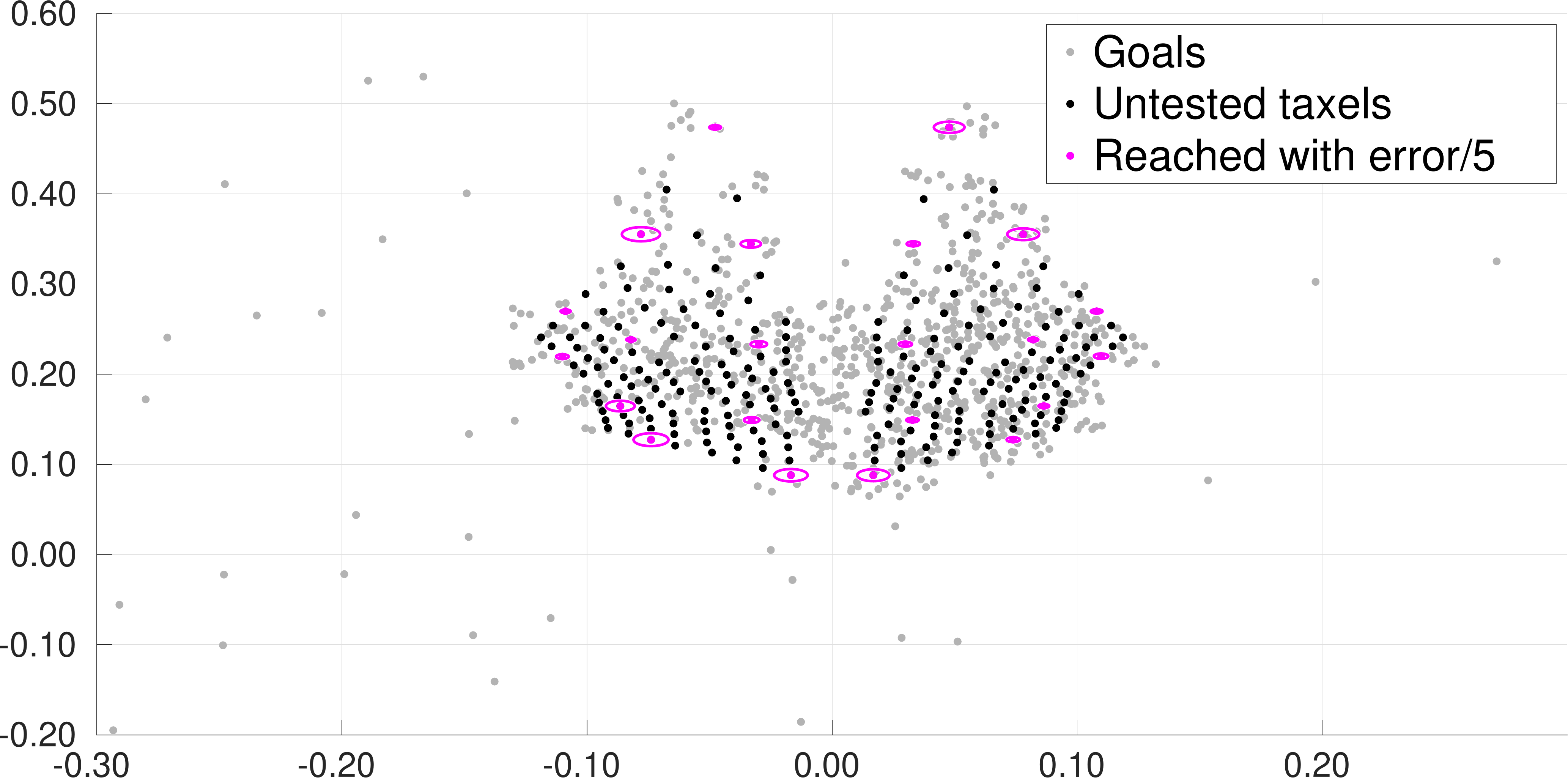}
  (f) DGB 32x32--high res.
\end{subfigure}

\caption{Right hand reaching for head. Low-res. skin (a and b); high-res. (c to f) -- observation space. See Fig.~\ref{fig:proj-lowres-torso} for details.}
\label{fig:proj-lowres-head}
\end{figure}

The projections (Fig.~\ref{fig:proj-lowres-head}, a and b) show results consistent with the low res. torso. DGB is better than RGB, with average errors lower than the distance to the closest taxel, indicating that most of the taxels are reached perfectly in a majority of the 10 trials. DGB's goal generation covers all taxels, with no empty spots, compared to low-res and high res. torso. DO did not show notable improvement in performance (plots not shown). Contrary to low-res. torso, no subset of taxels seems to always be perfectly reachable, but there is no taxel or set of taxels showing an overall significantly higher error than others.

\paragraph{High-resolution skin}

\begin{figure}[!ht]
\centering
\begin{subfigure}{.24\textwidth}
  \centering \includegraphics[scale=1, width=1\linewidth]{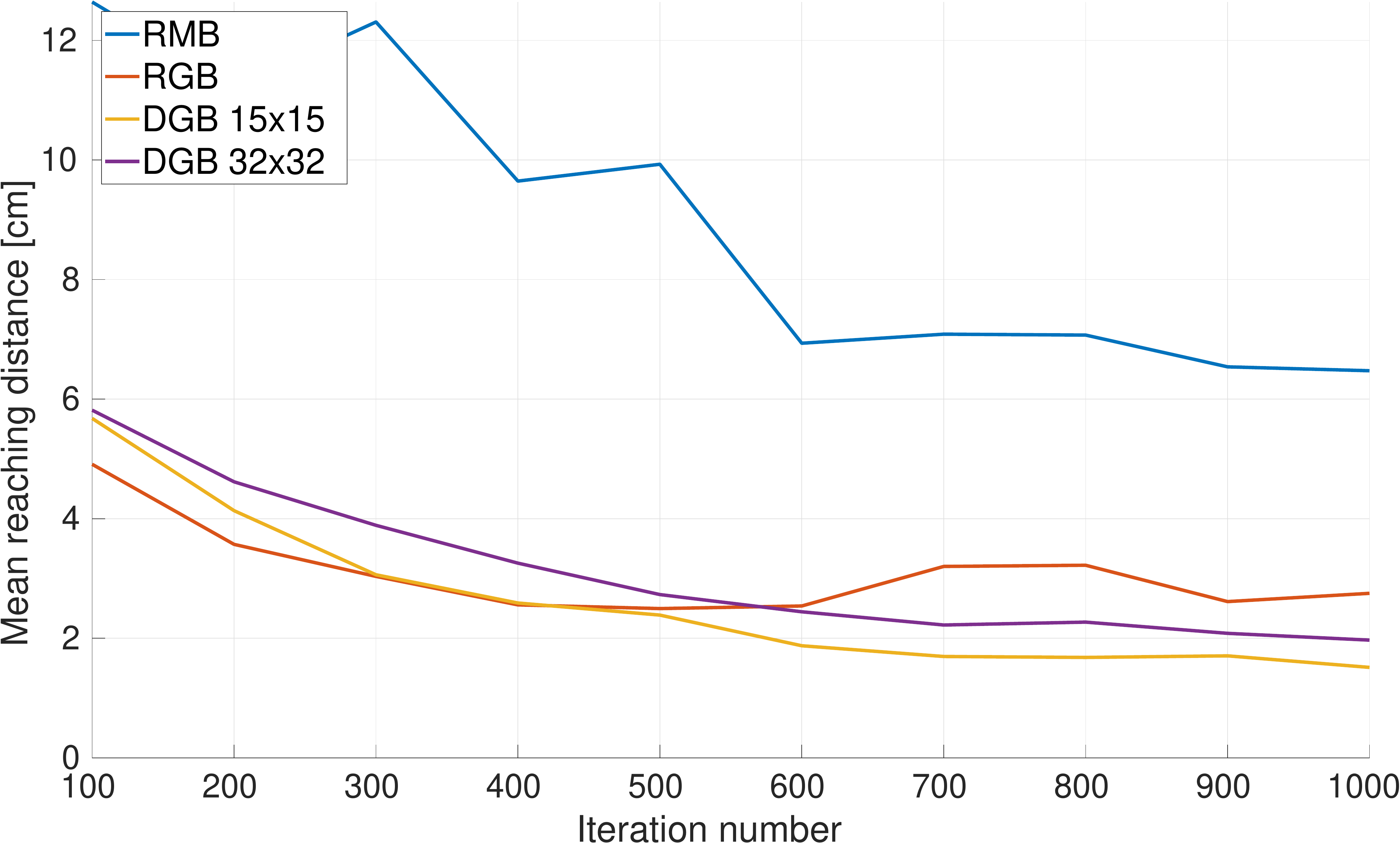}
\end{subfigure}%
\begin{subfigure}{.24\textwidth}
  \centering \includegraphics[scale=1, width=1\linewidth]{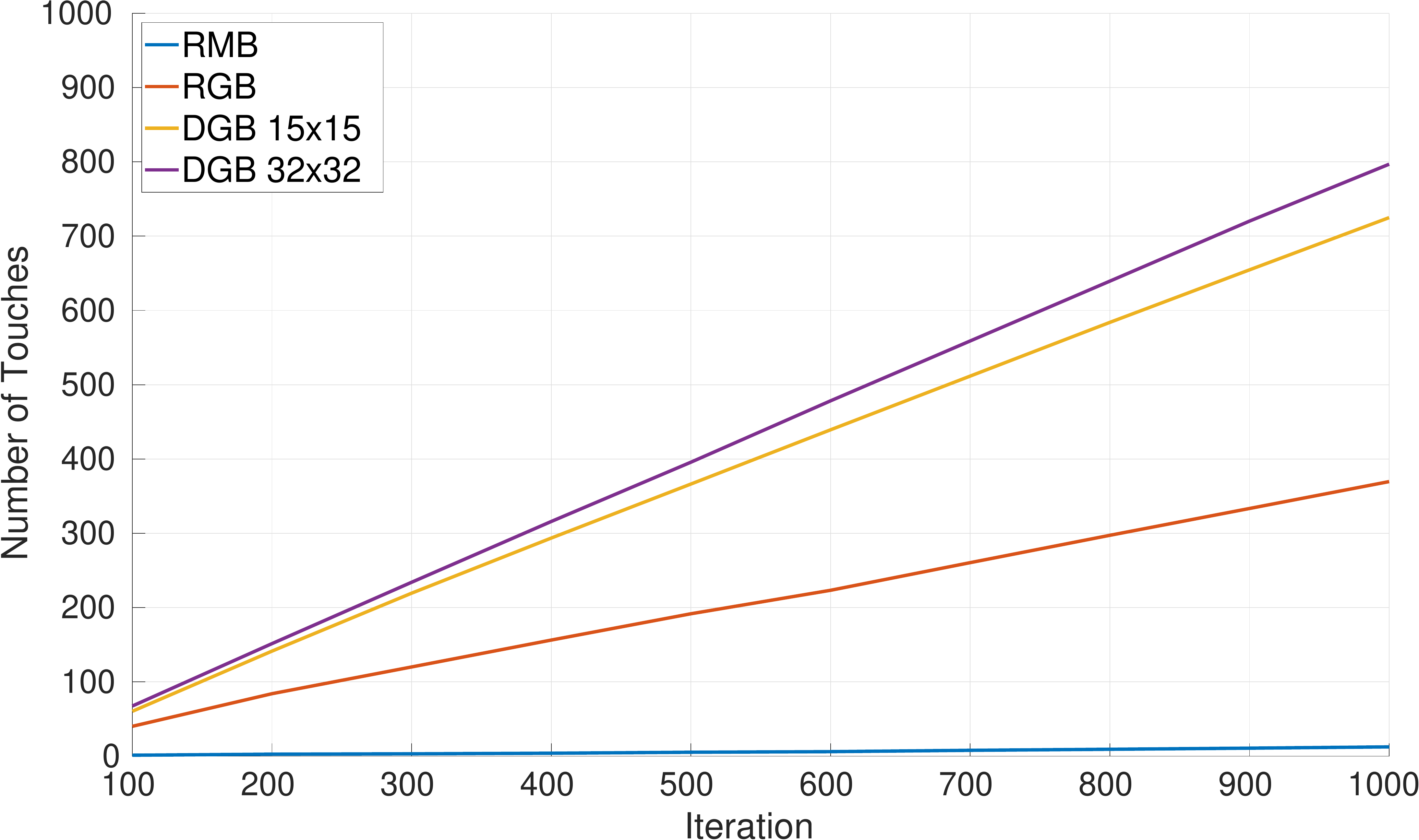}
\end{subfigure}
\caption{Right hand reaching for head, high-res. skin. (Left) MRE. (Right) Number of touches generated.}
\label{fig:results-highreshead}
\end{figure}

The results for high resolution (Fig.~\ref{fig:results-highreshead}) show the same change observed between high-res. torso and low-res torso, with a clear decrease of MRE over the course of the trials for DGB. However, MRE is higher than for the low-res. head, likely due to the cylindrical projection distortions that increase the distance between the taxels on the upper part of the head.
%Compared to low-res. skin, mean reaching error (Fig.~\ref{fig:results-highreshead}, left) starts at a much higher value, this is probably due to the projection of the top parts of the head, that end up farther from the other taxels, increasing the error if these taxels where not reached. However, the end results, for both for MRE and the number of touches (Fig.~\ref{fig:results-highreshead}, right), are very close to the low res. skin.
Looking at the observation space (Fig.~\ref{fig:proj-lowres-head}, c to f), contrary to high-res. torso skin, RMB reaches some taxels, but with high errors. RGB shows several unreached taxels, mostly on the upper-left of the head. Both DGB show the lowest errors and similar results, with again no unreached taxels.

\section{Summary and Discussion of Experiments}
\label{sec:conclusion}
As would be expected, the most successful algorithms in learning to reach to the body in our experiments were those with a competence-based algorithm (discretized progress): monitoring the learning progress in different areas of the goal space (the skin surfaces). However, in our case, the results are less clear-cut compared to standard learning inverse kinematics (e.g., \cite{Baranes2009,Rolf2010}): goal babbling with or without discretized progress or direct optimization achieves similar performance. This has mostly to do with the measurement of the performance available to the learning algorithm and also during testing: if no contact on the skin is generated, no error is available and learning as well as ``external evaluation'' are compromised. For this reason, we complemented the results presentation by reaching error (MRE) with the number of touches achieved and the goal space projection. None of these provide a complete picture, but together they improve our understanding. Reaching for the head was overall more successful than reaching for the torso, presumably because the two additional joints on the neck were recruited.

\section{General Discussion and Future Work}
\label{sec:discussion}

First, the issue of motor redundancy should be discussed. The inverse model---from skin space to joint space---was learned directly from the training samples and it was best represented with the nearest neighbor algorithm. While \textit{direct inverse modeling} \cite{Jordan1992} is prone to the ill-posedness of the general inverse kinematics problem and the averaging over non-convex solutions sets, our solution circumvents this by performing the exploration in the goal space: alternative solutions exploiting motor redundancy are thus not sought. Additionally, in the nearest neighbor algorithm, no averaging takes place. However, there are some trade-offs associated with this choice. The solution found will in our case be the first solution found; it may thus depend on initialization or chance and may not be the best solution. \textit{Distal learning}, or \textit{learning with a distal teacher} \cite{Jordan1992}, as opposed to direct inverse modeling, is more versatile in that it allows the incorporation of additional constraints to channel the search for the (single) solution. However, while initially a single solution to a reaching target on the body may suffice, we know that we, adults, are capable of alternative solutions depending on context.
%nonconvexity: Jordan & Rumelhart 1992, Fig. 8, pg. 315-316; or \cite{Rolf2010}
% \item \cite{Jordan1992} Distal learning of an inverse model: avoids problem of averaging over multiple solutions and is goal-directed (similar to goal babbling?). ``In cases in which the forward mapping is many-to-one, the distal learning procedure finds a particular inverse model. Without additional information about the particular structure of the input-to-action mapping, there is no way of predicting which of the possibly infinite set of inverse models the procedure will find. As is discussed later, however, the procedure can also be constrained to find particular inverse models with certain desired properties. ''
Distal learning allows the incorporation of a forward model and inverse model in series. Such a solution is more versatile in that the forward model, which is unique, ``disambiguates'' between alternatives coming from the, one-to-many, inverse mappings and can check their correctness. Human motor control in the cerebellum may be employing multiple paired forward and inverse models \cite{Kawato1999} (see the MOSAIC model \cite{Haruno2001}). Distal learning can thus in principle deal with a redundant system, but the problem is that the motor error is not directly observable \cite{Rolf2010}. A solution that would allow the agent to find one solution for every reaching target first, but add and keep alternatives later on, remains our future work. The mixed---composite forward-inverse models---can be a solution (see \cite{Nguyen-Tuong2011} for a survey).
% Handling of redundancy; goal babbling deliberately avoids it / ``deals'' with it - learns direct inverse function with one solution; \cite{Rolf2010} claims that this is developmentally plausible - let's verify that
It is also worth considering how the task studied here---reaching to the body---differs from reaching in general. Self-touch configurations are more kinematically constrained than reaching in free space in front of the body and hence the effective motor redundancy is likely lower. This is even more the case for the experiments used here in which only five DoF of the Nao arm were employed. 
%we should test physically or analyze... get the manipulability visualization like cive for iCub?

%    \item  learning body - skin space - similar to general IK learning, but confined to a manifold in 3D; we have actual skin activations rather than Cartesian position and error 
Second, the use of nearest neighbor algorithm for the inverse model representation has to be discussed. It has the following advantages: (i) incremental learning is simple and requires registering pairs from input and goal space only (``lazy learning''); (ii) there is no averaging or interpolation of samples (avoiding the problem of non-convexity of the solution space). The disadvantages are: (i) computational complexity: all experience is stored in memory and upon retrieval---query to the inverse model---time is required to find the nearest neighbor; (ii) susceptibility to noise: in our scenario, ``phantom'' skin activation would be catalogued together with the current joint configuration and contaminate the model. (iii) mapping will not be smooth: adjoining skin receptors will not necessarily map to nearby joint configurations. 
Baranes and Oudeyer~\cite{baranes_2013} deem nonparametric methods (like nearest neighbor) suitable for their problems (including inverse kinematics) and the complexity problem can be mitigated by efficient implementation \cite{Muja2009}. Alternative representations of the inverse model could be local regression methods (e.g, Locally Weighted Projection Regression; Sigaud et al.~\cite{Sigaud2011} for a survey). How such mappings are encoded by the brain is an open question.

Third, the representations of the input and output spaces importantly influence what can be learned and how. Regarding the input, motor, space, we have discussed some alternatives from \cite{baranes_2013,mannella_2018} in Section~\ref{sec:rel_work}. In our representation, the actual execution of the movement---initiation, termination, and its dynamics---has not been addressed  and such separation of movement preparation and control may not be justified~\cite{Schoner2018}. Mannella et al.~\cite{mannella_2018} do consider this aspect and observe for example that easy postures are acquired before hard ones. Dynamic Movement Primitives \cite{Ijspeert2003} seem a good candidate for such representation.
%The computational perspective of Cisek could be used as a starting point to add this important dimension~\cite{Cisek2001}.  
Regarding the ``skin space'', one could come closer to the biological reality by mimicking the non-uniform density of receptors (as done in \cite{Mori2010,mannella_2018}). On the representation level, self-organizing maps seem like a natural candidate \cite{mannella_2018,HoffmannStraka2018}. The distance metric required for the exploration will then be distorted as is typical for homuncular representations and present an additional challenge. Finally, the motor and sensory spaces could be treated in a more integrated manner as proposed by Marcel et al.~\cite{Marcel2016} who present a mathematical analysis of building a sensorimotor representation of a naive agent’s tactile space. 

%Discussion
%\begin{itemize}
    %\item see also Discussion (5.2) in Max's thesis 
    %\item Considering the limitations of what we are doing, the use of the simulator has limitations concerning the touch feedback we get. Sometimes, no touch is generated because it touches the plastic casing, even though it is very close to a taxel. This would not occur on the physical robot nor in humans, because of the fabric/skin, which deformations activate nearby taxels. With the fabric, it is also important/necessary to use the intensity of the activation of the taxel, while in the simulator, we currently use a binary activation value 
    %\item touch could be important for learning to reach -- see refs in \cite{Nguyen_ICDL_2019}
    %\item (ipsilateral reaches...)
    %\item what we did here was like imagining the child would self-generate goals to touch the skin. Do infants do that? Alternatives? RL framework - reward for skin touch - what would be the consequences? would be insufficient. reach one spot and stop.
%\end{itemize}

%Future work
%\begin{itemize}
 %   \item symmetry based exploration \cite{Rayyes2018}
%\end{itemize}

It is our ultimate goal to ground the model in biological data. The work of Schlesinger~\cite{Schlesinger2013} is an example investigating looking patterns of infants. In our scenario, there are two concrete ways how we plan to proceed. First, in our study, the robot is learning an inverse model: which motor commands to use to reach to targets on its body. The performance for different body parts and at different stages of development can be compared with behavioral data from infants reaching for vibrotactile stimuli on the body \cite{Leed2019}. For example, we should analyze how infants deal with the redundancy of their motor system in this particular case: during different ``stages'' in their development, do they use the same or distinct configurations to reach for targets on the body? If the latter were the case, the goal exploration strategies that suppress the redundancy of the motor system may not be appropriate. Also, with different initial postures, do infants tend to go to a canonical posture first? There is evidence suggesting that this may be the case in infants \cite{Berthier1999} and adults learning a new task \cite{Rohde2019}.
%or Graziano - monkey - no matter where you start...
Second, statistics obtained from studies observing spontaneous touches to the body in infants \cite{DiMercurio2018,Thomas2015}---such as how often infants touch particular body parts, in which sequence etc.---could be fed into the robot simulator to train the inverse model and the results in terms of reaching performance to targets on the body compared with those obtained from the computational exploration strategies. Alternatively, we could aim to model the exploration process itself and obtain similar self-touch statistics as an emergent property. Discovering signatures of curiosity-driven learning in the brain is an active research area~\cite{Gottlieb2018}.
%, employing fMRI~\cite{Gruber2014} or EEG and body states~\cite{Appriou2019}.
Only behavioral data poses a greater challenge. With carefully designed experiments, one may be able to discern which cost function the ``learning machine'' is using~\cite{Cashaback2017}. Discriminating spontaneous vs. systematic exploration in naturalistic observations (like \cite{DiMercurio2018,Thomas2015}) remains to our knowledge an open question.

%\item Rohde et al.~\cite{Rohde2019} state that the features of goal babbling are that ``goal-irrelevant parts of the motor space or parts that contain solutions that are redundant to the initially learned one are thus never explored.'' While the former may not be problematic, the latter seems like a serious intrinsic limitation. Humans certainly can exploit the redundancy of the motor system when needed. 

\section*{Acknowledgment}
 This work was supported by the Czech Science Foundation (GA CR), project EXPRO (nr. 20-24186X). We acknowledge feedback from discussions with Pierre-Yves Oudeyer, Clement Moulin-Frier, Igor Farka\v{s}, and Gianluca Baldassarre at earlier stages of the project. Karel Zimmermann co-supervised the thesis \cite{Shcherban2019} that was the foundation of this work. Martin J\'{i}lek developed an earlier version of the simulation environment.

%\section*{References}

\bibliographystyle{./bibliography/IEEEtran}
%\bibliography{./bibliography/IEEEabrv,./bibliography/IEEEexample}
\bibliography{./bibliography/ICDL2020_NaoSelfExploration}

\end{document}